\documentclass[10pt,twocolumn,letterpaper]{article}
\usepackage{todonotes}

\usepackage{iccv} 
\definecolor{iccvblue}{rgb}{0.21,0.49,0.74}

\usepackage{tikz}

\usepackage{algorithm}
\usepackage{algpseudocode}
\usepackage[commentColor=black,noEnd=false]{algpseudocodex}
\usepackage{pifont}
\usepackage{amssymb}

\newcommand{\xmark}{\ding{55}}  
\newcommand{\cmark}{\ding{51}}
\usepackage{lipsum}
\usepackage{subcaption}
\usepackage{tikz}
\usepackage{booktabs}
\usepackage{multirow}
\usepackage{float}
\usepackage{dsfont}
\usepackage{algorithm}
\usepackage[commentColor=black,noEnd=false]{algpseudocodex}
\usepackage{pifont}
\renewcommand{\cmark}{\text{\ding{51}}}
\renewcommand{\xmark}{\text{\ding{55}}}

\definecolor{figred}{HTML}{B85450}
\definecolor{figgreen}{HTML}{82B366}
\definecolor{figblue}{HTML}{6C8EBF}
\definecolor{figyellow}{HTML}{D6B656}
\definecolor{figpurple}{HTML}{785C85}

\usepackage[accsupp]{axessibility} 

\usepackage[pagebackref,breaklinks,colorlinks,allcolors=iccvblue]{hyperref}
\PassOptionsToPackage{table,svgnames,dvipsnames}{xcolor}
\usepackage{xcolor}
\usepackage{colortbl}

\usepackage{textcomp}
 
\usepackage[inkscapelatex=false]{svg}
\usepackage{graphicx}
 \usepackage{multirow} 
\usepackage{subcaption}

\usepackage{bm}
\def\paperID{10044} 
\def\confName{ICCV}
\def\confYear{2025}
    
\title{Decoupling Continual Semantic Segmentation }

\author{Yifu Guo\textsuperscript{1,2,*},
        Yuquan Lu\textsuperscript{1,2,*}, 
        Wentao Zhang\textsuperscript{1},
        Zishan Xu\textsuperscript{2},
        Dexia Chen\textsuperscript{1},\\
        Siyu Zhang\textsuperscript{3},
        Yizhe Zhang\textsuperscript{4},
        Ruixuan Wang\textsuperscript{1,†}\\
        \\
        \textsuperscript{1}Sun Yat-sen University, 
        \textsuperscript{2}Shanghai Jiao Tong University,\\
        \textsuperscript{3}Southwest University, 
        \textsuperscript{4}University of Notre Dame\\
        {\tt\small \{20223801024\}@m.scnu.edu.cn, \{wangruix5\}@mail.sysu.edu.cn}
          \thanks{Equal contribution.\\ †Corresponding author.}
       }
\begin{document}
\maketitle
\begin{abstract}
Continual Semantic Segmentation (CSS) requires learning new classes without forgetting previously acquired knowledge, addressing the fundamental challenge of catastrophic forgetting in dense prediction tasks.  However, existing CSS methods typically employ single-stage encoder-decoder architectures where segmentation masks and class labels are tightly coupled, leading to interference between old and new class learning and suboptimal retention-plasticity balance. We introduce \textbf{DecoupleCSS}, a novel two-stage framework for CSS.  By decoupling class-aware detection from class-agnostic segmentation, 
DecoupleCSS enables more effective continual learning, preserving past knowledge while learning new classes. The first stage leverages pre-trained text and image encoders, adapted using LoRA, 
to encode class-specific information and generate location-aware prompts. In the second stage, the Segment Anything Model (SAM) is employed to produce precise segmentation masks, ensuring that segmentation knowledge is shared across both new and previous classes. This approach improves the balance between retention and adaptability in CSS, achieving state-of-the-art performance across a variety of challenging tasks. Our code is publicly available at:\href{https://github.com/euyis1019/Decoupling-Continual-Semantic-Segmentation}{https://github.com/euyis1019/Decoupling-Continual-Semantic-Segmentation}.
\end{abstract}
\vspace{-5mm}
\section{Introduction}
\label{sec:intro}

Continual Semantic Segmentation (CSS) addresses a practical scenario where new segmentation tasks with novel classes emerge over time~\cite{cermelliModelingBackgroundIncremental2020,douillardPLOPLearningForgetting2021,2024SurveyCISS}. A machine learning model must effectively learn these new classes while retaining previously learned old knowledge, ensuring that old knowledge is not forgotten. CSS has many real-world applications, such as autonomous driving~\cite{Lidarss}, medical imaging~\cite{mediccss,IncreNucleiSegmentation} and remote sensing~\cite{remotecss}. As a dense prediction task, CSS is particularly susceptible to semantic shifts, where the model's understanding of class relationships may evolve over time. Furthermore, similar to other continual learning tasks, CSS faces the challenge of catastrophic forgetting, where previously learned knowledge can be influenced and overwritten by new information.

    
    

\begin{figure}[t]
    \centering
    \includegraphics[width=\linewidth]{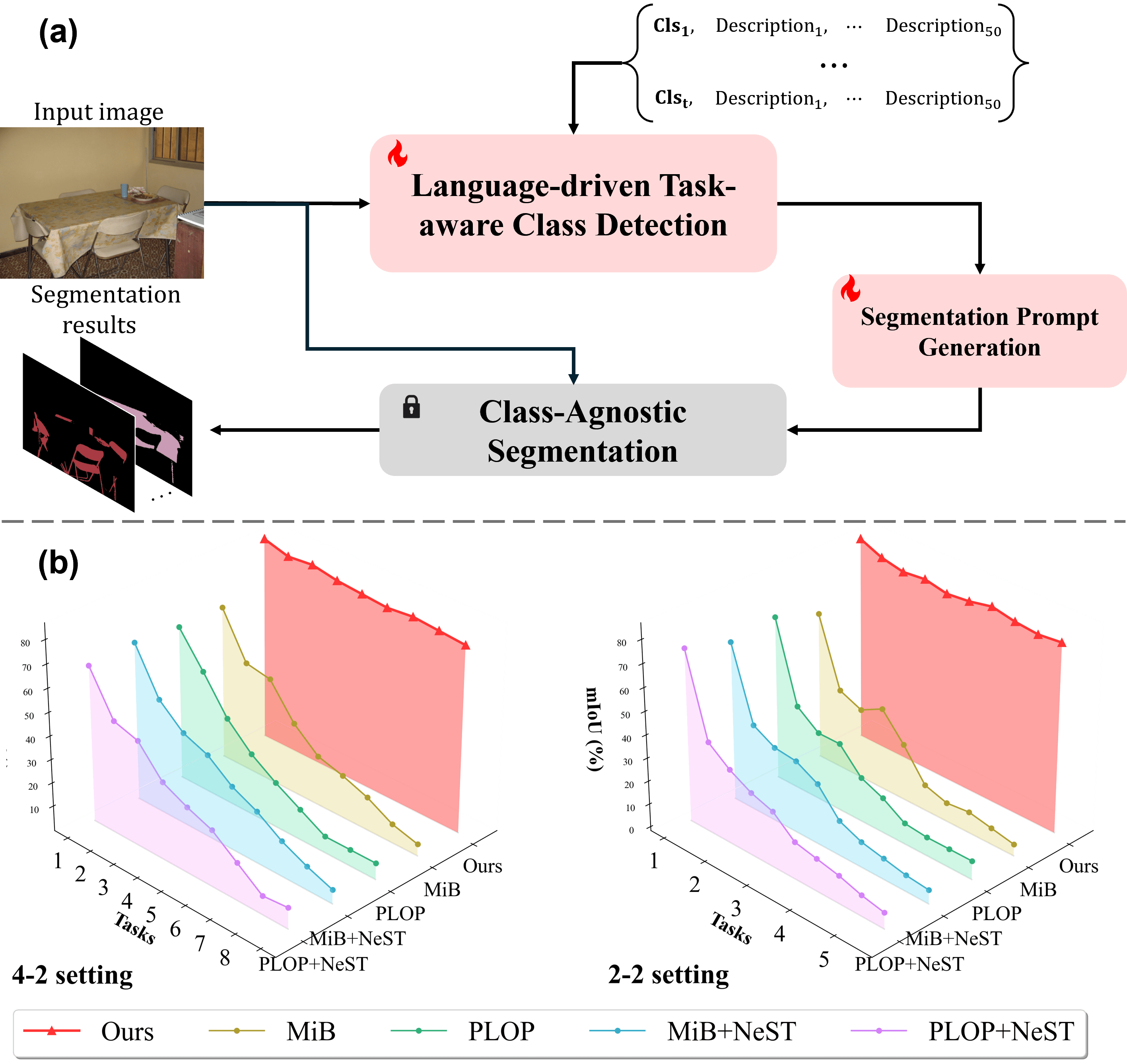}
    \label{fig:intro_represatitive_setting}
    \caption{\textbf{Overview of the proposed method.} (a) The overall architecture. 
(b) Representative results on challenging settings (2-2 and 4-2) for CSS on Pascal VOC 2012.}
\label{fig:combined}
\end{figure}
Several methods have been proposed to address the challenges in CSS. Data replay~\cite{recall,2023datareplay} is effective but requires storing raw data/features, leading to linear memory growth and potential privacy concerns. 
Regularization techniques~\cite{CVPR2020ModelingTB} penalize deviations in important model parameters for old classes, but 
often struggle to balance retention and plasticity. Pseudo-labeling~\cite{douillardPLOPLearningForgetting2021,pseudo2019} and knowledge distillation~\cite{su2024Distillation,wang2024Distillation,yang2023distillation} preserve old class knowledge by generating pseudo-labels or distilling knowledge from the old model, but still can cause background class drift when uncertain pixels are labeled as background. To mitigate this, the background weight transfer method~\cite{ECCV2024MitigatingBS, ipseg} initializes new class classifiers with background weights. However, it may mix new class features with background features, especially in cases of limited training data. 


Most prior CSS methods employ a single-stage (encoder-decoder) segmentation paradigm, as seen in models such as DeepLab-V3~\cite{chen2017rethinking}.  In these methods based on pixel-level multi-class classification, 
segmentation masks (intra-class consistency) and class labels (inter-class discrimination) are tightly coupled within a shared set of model parameters. This creates a significant challenge in the continual learning setting where the supervisory signals needed for inter-class discrimination between old and new categories are systematically removed, which partly causes catastrophic forgetting.

To tackle the CSS challenge, we propose a two-stage segmentation framework called DecoupleCSS 
in which class-aware perception (Figure~\ref{fig:combined}a, red components as the first stage) is decoupled from Class-Agnostic Segmentation (CAS, gray component in Figure~\ref{fig:combined}a as the second stage), enabling a separation of concerns where continual learning targets the class-aware detection phase, while the segmentation module can be shared across old and new tasks. 
In the first stage, class-specific semantic textual information is 
used to guide an adapted image encoder to extract class-relevant features. Note that different tasks share the same pre-trained frozen image encoder but have their own task-specific visual adapters which are optimized during continual learning.
Such class-relevant visual features are then used to detect existence of classes in the input image and to 
generate class-aware and location-specific prompts. In the second stage, these prompts activate SAM to generates precise segmentation masks. Our contributions are summarized below.
\begin{itemize}
\item At the framework level, this study presents a novel perspective and approach for Continual Semantic Segmentation (CSS). We advocate using class-agnostic foundation models (e.g., SAM) as a cornerstone for practical CSS research and propose a separation of class-aware and class-agnostic components in CSS learning. This explicit decoupling enables effective, focused continual learning for detection (class-aware), while segmentation knowledge (class-agnostic) is shared across classes.

\item At the method level, 
a novel task-specific class detection strategy and a novel class-specific prompt generation strategy are proposed to employ SAM 
for both new and previous tasks. Our method effectively integrates new classes while preserving prior knowledge. 

\item Our method effectively balances old knowledge retention with new knowledge learning and yields significantly improved accuracy and adaptability, demonstrating superior CSS performance 
across a diverse range of challenging CSS tasks (see representative results in Figure~\ref{fig:combined}b). 
\end{itemize}
\section{Related Work}
\label{sec:related-work}
\noindent\textbf{Continual Semantic Segmentation} The challenges in CSS include catastrophic forgetting, stability-plasticity dilemma, and semantic (background) shift~\cite{2024SurveyCISS}. PLOP~\cite{douillardPLOPLearningForgetting2021} employs pseudo-labels generated by the prior model and multi-scale local distillation to preserve old knowledge. A series of works~\cite{cermelliCoMFormerContinualLearning2023, ECCV2024MitigatingBS,shang2023distillation,yang2023distillation, InheritwithDistill,zhao2023distillation} follow this way, aiming to enhance model stability by reducing information loss and preserving existing knowledge.
For example, BalConpas~\cite{ECCV2024StrikeStrikeaBalanceinContinualPanopticSegmentation} selectively distills the most relevant feature, and Cs$^2$KCA~\cite{ECCV2024Cs2KCA} uses pixel-level features as a prototype for each class. 
Though Exemplar Replay~\cite{CVPR2022CL,yan2021DER} has advanced significantly in CSS from three aspects: Sample Replay~\cite{samrep2021ssul,samrep2022pin,samrep2021RECALL,samrep2017iCaRL,samrep2024tikp}, Feature Replay~\cite{ferep2022dynamic,ferep2022,ferep2022semi,ferep2020} and Auxiliary Data~\cite{AD2021,AD2023foundation}, dynamic networks~\cite{DA2022decomposed,DA2022continual,DA2023fairness,DA2023effects,DA2023endpoints,DA2022deep,DA2022learning} stand out for their efficacy in preserving crucial parameters and enabling flexible task-specific adjustments~\cite{CoMasTRe2024}, making them particularly advantageous in CSS. While recent CSS methods like CoMasTRe~\cite{CoMasTRe2024} and CoMFormer~\cite{cermelliCoMFormerContinualLearning2023} based on Mask2Former~\cite{Mask2Former} also explore decoupling strategies, with CoMasTRe focusing on objectness learning and classification separately, our framework follows a distinct detection-then-segmentation paradigm which shows superior performance.

\noindent\textbf{Language-Driven Continual Learning}  Inspired by the observation that humans effectively acquire new visual knowledge through language and motivated by the broad applications of pre-trained vision-language models (VLMs)~\cite{Huang_Huang_Zhang_Tian_Feng_Zhang_Xie_Li_Zhang_2023,Radford_Kim_Hallacy_Ramesh_Goh_Agarwal_Sastry_Amanda_Mishkin_Clark_et_al._2021}, previous studies have explored VLMs in continual learning classification tasks~\cite{Knowledge-Aware,wentaozhang2024continual,zheng2024large}. However, applications of VLMs for Continual Semantic Segmentation (CSS) are much less explored, primarily because the dense annotations required for segmentation cannot be directly leveraged by these models; only a few studies have explored their potential in Weakly Supervised Continual Semantic Segmentation~\cite{FoundationModelDWISS}. To our best knowledge, our work is the first to leverage VLMs in dense-annotated CSS.
\section{Method}
\label{sec:method}
This study focuses on class-incremental semantic segmentation (CISS), where a model learns a sequence of \(T\) semantic segmentation tasks. In the  \(t\)-th task (\(t=1, 2, \ldots, T\)), the model is updated to learn to segment each image of the task into regions of \(c_t\) new foreground classes and a background class, while preserving the ability to segment regions corresponding to the previously learned \(c_1 + c_2 + \ldots + c_{t-1}\) classes. In each training image of task \(t\), the background may contain objects of both previously learned foreground classes (from tasks \(1\) to \(t-1\)) and future classes (from tasks \(t+1\) to \(T\)); this overlapped setting in CISS is more challenging than the disjoint scenario where future classes are absent during learning task $t$. After learning task \(t\), the model is expected to segment any test image containing any subset of the \(c_1 + c_2 + \ldots + c_t\) learned classes.




\subsection{Framework Overview}
\begin{figure}[t]
    \includegraphics[width=1.0\linewidth]{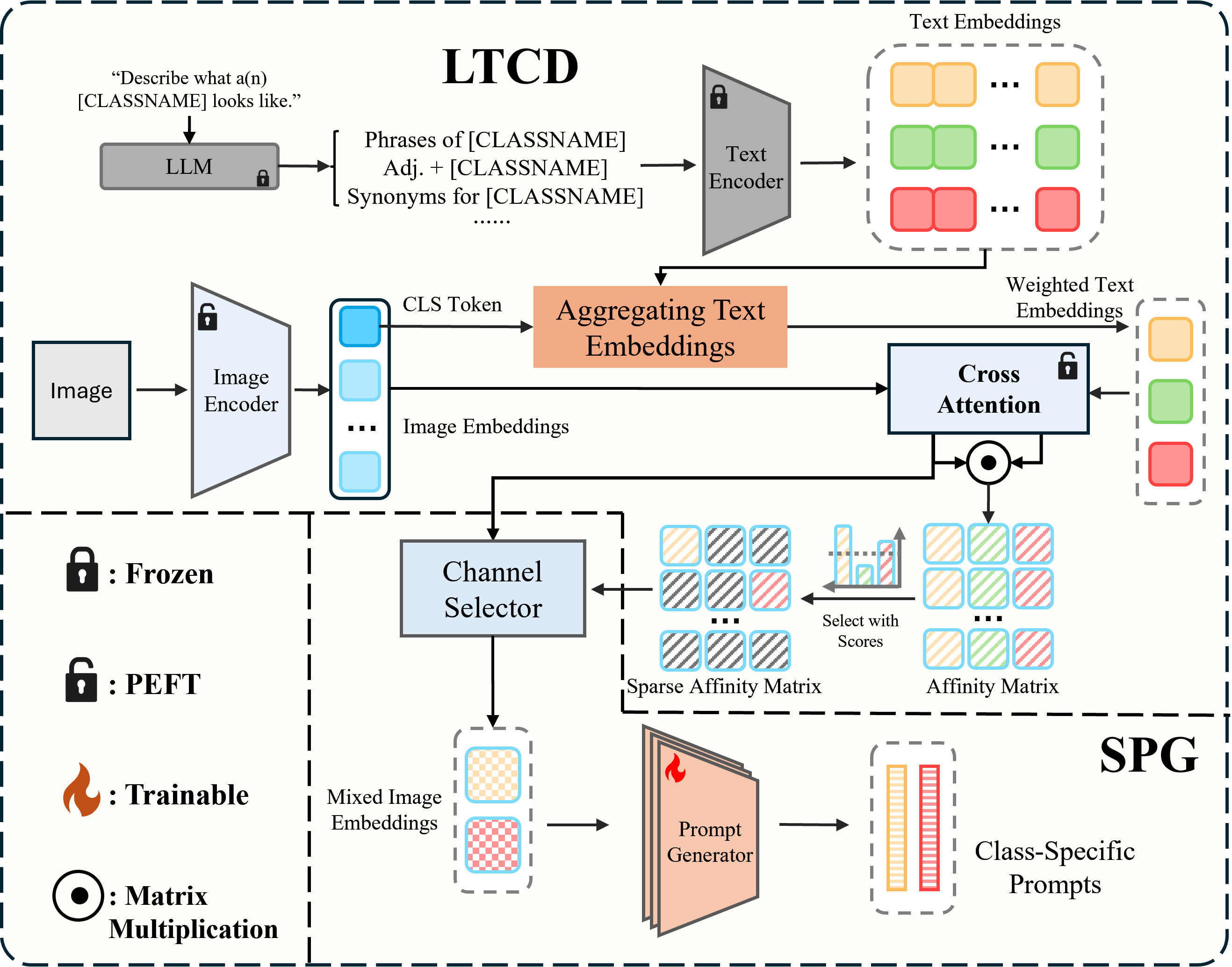}
    \vspace{-5mm}
    \caption{\textbf{Workflow of the LTCD and SPG module.} 
    }
    \label{fig:TCP}
\end{figure}
\vspace{-1.2mm}

The proposed CSS framework (Figure~\ref{fig:TCP}) is developed based on the insight that the process of segmenting an image can be decoupled into two independent stages, the existence detection of foreground classes in the image (Stage-I) and the accurate delineation of the foreground region(s) for each detected class (Stage-II). With more and more segmentation tasks learned, Stage-I should be able to detect existence of more and more classes. To achieve this goal,  a novel language-driven and task-aware class detection (LTCD) module built on pre-trained large language and vision models is designed. This module can find existence of classes for each learned task with the help of task-specific adapters and a detection, and thus catastrophic forgetting of old knowledge for class detection is avoided.

For Stage-II, inspired by the class-agnostic segmentation capability of the recently developed foundation segmentation model SAM, we employ the pre-trained SAM to automatically annotate the region of each detected class. The challenge here is to provide location-relevant prompts informing SAM where to segment for each detected class. To solve this challenge, a segmentation prompt generation (SPG) module is designed to activate SAM for accurate region segmentation of each detected class. The SPG module bridges the LTCD and the CAS modules, taking the patch-level visual embeddings and language-guided class-wise image embeddings from the LTCD module as input. 

In this framework, task-wise continual learning in the LTCD module and class-wise continual learning in the SPG module are responsible to continually learn  knowledge of new classes, while the class-agnostic CAS module is responsible for accurate region annotation given location-relevant prompts. 
By isolating class detection from class-agnostic region annotation, our method provides a robust, modular solution for continual semantic segmentation. 

\subsection{Language-driven Task-aware Class Detection}
\label{subsec:stage-1}

Suppose the model will be learning the $t$-th task including $c_t$ new classes. The LTCD  module utilizes textual prior knowledge of each class in task $t$ to help adapt the pre-trained image encoder and the cross attention part for task $t$. In particular, the encoded textual prior is used in the text-image cross attention blocks to help extract the class-specific visual embedding from the adapted image encoder for rough class awareness. Note that only the task-specific adapters which are added to the pre-trained image encoder and the pre-trained text-image cross attention block are learnable. 

\subsubsection{Text Encoding} \label{sec:generationAnchors}

For each of the \(c_t\) classes, a number of \(M+1\) descriptive phrases are generated (see \Cref{sec:prompt} for the complete template and process). Among them, \(M\) phrases are generated by the large language model~\cite{OpenAI_2023} using a pre-defined prompt~\cite{Bai_Xia_2023,Pratt_Liu_Farhadi_2022}. Unlike conventional approaches that generate complete sentences with redundant linguistic elements, our method employs concise phrasal descriptions (e.g., adjective + {class name}) to capture essential visual attributes. This focused representation enhances feature distinctiveness and improves class discrimination capabilities. Each phrase is then sent to a pre-trained text encoder to obtain the corresponding text embedding. Consequently, for the \(k\)-th class (\(k=1, 2, \ldots, c_t\)), a set of \(M+1\) text embeddings \(\{\mathbf{g}_{k,1}, \mathbf{g}_{k,2}, \ldots, \mathbf{g}_{k,M+1}\}\) are generated. The purpose of generating multiple versions of textual descriptions and corresponding text embeddings for each class is to increase diversity and enhance the robustness of the subsequent aggregation step.

\noindent\textit{Aggregating text embeddings. } While using multiple textual embeddings enriches class representations by providing a wealth of contextual information, it simultaneously introduces irrelevant or misaligned semantics that do not correspond to the specific content of the input image.
To mitigate this issue, we propose an adaptive re-weighting strategy that adjusts the influence of each text embedding based on its relevance to the visual content of the current input image. For an input image $\mathbf{x}_i$, the weight for each text embedding is
\begin{equation}
    \alpha_{i,k,m} = \frac{\exp(s_{i,k,m})}{\sum_{j=1}^{M+1} \exp(s_{i,k,j})} \,,
\end{equation}
where $m \in \{1, 2, \dots, M+1\}$, $k \in \{1, 2, \dots, c_t\}$, and the score $s_{i,k,j} = \cos (\mathbf{V}_{i}^{cls}, \mathbf{g}_{k,j})$ measures the cosine similarity between the \textit{cls} token of visual embedding $\mathbf{V}_{i}$ for the input image $\mathbf{x}_i$ from the adapted image encoder (see following subsection
) and the $j$-th text embedding $\mathbf{g}_{k,j}$ of class $k$. 
The final text embedding for class \(k\) is generated based on weighted sum of all $M+1$ text embeddings, \ie,
\begin{equation}
    \mathbf{e}_{i,k} = \sum_{m= 1}^{M+1} \alpha_{i,k,m} \cdot \mathbf{g}_{k,m} \,.
\end{equation}
Note that $\mathbf{e}_{i,k}$ is image-wise, \ie, different input images would lead to different text embeddings $\mathbf{e}_{i,k}$ for same class $k$.
The set of text embeddings $\mathbf{E}_i=[\mathbf{e}_{i,1}, \mathbf{e}_{i,2}, \ldots, \mathbf{e}_{i,c_t}]  \in \mathbb{R}^{c_t \times d}$ for the $c_t$ classes of task $t$ will be utilized along with the visual embedding of input image $\mathbf{x}_i$ (see below) for language-driven class detection.

\subsubsection{Task-Specific LoRAs}  \label{sec:vencoder2}

In our framework, we initialize the pre-trained image encoder (with a Swin Transformer backbone) and cross-attention module from Grounding DINO~\cite{liu2024groundingdinomarryingdino} which are endowed with robust feature extraction and cross-modal alignment capabilities. However, while these pre-trained components provide reliable general representations, they are not specifically optimized for the nuanced requirements of learning novel classes incrementally.


%
Here, task-specific LoRA~\cite{hu2021lora} adapters are added into both the image encoder and the cross-attention module to learn each new segmentation task. For the image encoder, these lightweight adapters are embedded in each self-attention layer, enhancing discriminative feature extraction for new classes while maintaining the output structure as $\mathbf{V}_i \in \mathbb{R}^{N \times d}$ for each input image $\mathbf{x}_i$, where $N$ represents visual tokens and $d$ is the embedding dimension.
In the cross-attention module, which coordinates bidirectional information flow between modalities,  LoRA adapters are incorporated in the linear projections that generate attention `keys' and `values'. This approach enables effective text-guided visual feature optimization through multiple interaction layers, producing enhanced representations $\mathbf{V}'_{i} \in \mathbb{R}^{N \times d}$ and $\mathbf{E}'_{i} \in \mathbb{R}^{c_t \times d}$ that better capture cross-modal relationships while maintaining computational efficiency.

Note that the learnable adapters are task-specific, which prevents negative transfer and catastrophic forgetting across incremental tasks. All learned adapters for previous tasks 1 to $t-1$ are preserved but not utilized when the model learns task $t$. However, during model inference, adapters of each learned task will be utilized to segment any test image. 

\subsubsection{Semantic Alignment and Token Selection}

This section details the process of establishing precise semantic correspondence between image regions and class concepts, a crucial step for class-aware segmentation. While the cross-attention module (described above) refines visual and textual embeddings ($\mathbf{V}'_{i}$ and $\mathbf{E}'_{i}$ respectively), an explicit alignment measure is needed to guide the segmentation process.

\noindent\textit{Affinity matrix construction. }To quantify the alignment between visual tokens and textual class embeddings, we compute an affinity matrix, $\mathbf{S}_i$, for each image $\mathbf{x}_i$. This matrix represents the pairwise similarity between each visual token and each class embedding, computed as follows
\begin{equation}
\mathbf{S}_i = \cos(\mathbf{V}'_{i} , {\mathbf{E}'_{i}}^T) \in \mathbb{R}^{N \times c_t} \,,
\end{equation}
where \(\cos(\cdot, \cdot)\) denotes the cosine similarity measurement. The resulting $\mathbf{S}_i$ is a matrix where each element $\mathbf{S}_i[n, k]$ represents the alignment score (dot product) between the $n$-th visual token and the $k$-th class embedding.  Higher values indicate stronger alignment.


\noindent\textit{Thresholding for salient token selection. } Some image regions may be irrelevant to learned classes, e.g., background regions in images.  To filter out these non-salient regions and focus on the most relevant visual tokens, we apply a thresholding operation to the affinity matrix $\mathbf{S}_i$, creating a sparse affinity matrix $\mathbf{S}'_i$
\begin{equation}
\mathbf{S}'_{i}[n,k]=
\begin{cases}
\mathbf{S}_i[n,k], & \quad \text{if } \mathbf{S}_i[n,k] \geq \tau \\
0, & \quad \text{otherwise}
\end{cases}
\end{equation}
where $\tau$ is a pre-defined threshold.  This operation effectively eliminates weak associations, retaining only visual tokens with strong alignment to at least one class. This sparsity reduces noise and computational cost in subsequent steps.

\noindent\textit{Selection of semantically aligned tokens.} From the sparse affinity matrix $\mathbf{S}'_i$, we select the visual tokens that exhibit strong alignment with any class.  The selection process is defined as follows:

\noindent\underline{Identify relevant rows:} We identify the rows in $\mathbf{S}'_i$ that contain at least one nonzero element. The collection of these rows corresponds to visual tokens of interest, denoted  $\mathcal{I}_{row}$.  


\noindent\underline{Extract selected tokens:} We extract the visual embeddings corresponding to the selected row indices. This forms the set of selected visual embeddings: \(\mathbf{V}_i^{\text{sel}} = \{\mathbf{V}'_i[n,:] | n \in \mathcal{I}_{row}\}\).

\noindent\underline{Determine semantic associations:} For each selected token in $\mathbf{V}_i^{\text{sel}}$, we determine its associated class by identifying the class with the highest alignment score in the corresponding row of the sparse affinity matrix $\mathbf{S}'_i$.  This is represented by the set of  index pairs in the form of (\texttt{selected token, class}), i.e., \(\mathbf{C}_i = \{(n, k) | n \in \mathcal{I}_{row}, k = \operatorname*{arg\,max}_j\mathbf{S}'_i[n, j]\}\).
\begin{figure*}[t]
    \centering
    \includegraphics[width=\textwidth]{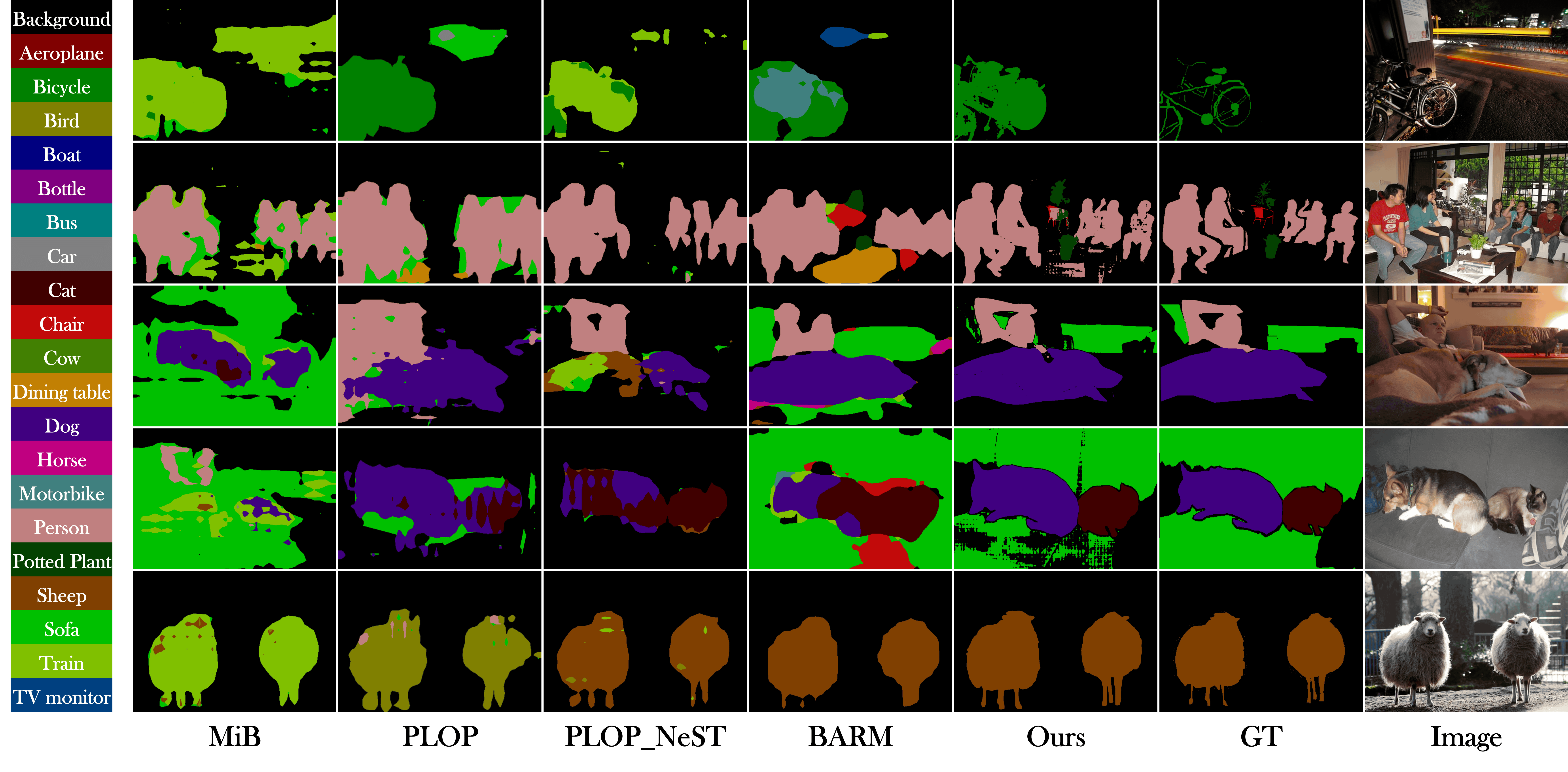}
    \caption{The visualization comparison from the last task on the Pascal VOC2012 10-1 setting.}
    \label{fig:main_results}
\end{figure*}

\vspace{-2mm}
\subsection{Class-Specific Prompt Generation}
\label{sec:spg}

This section describes the generation of class-specific positional prompts from the selected visual embedding ($\mathbf{V}_i^{\text{sel}}$), which are then used for class-agnostic segmentation.  
We propose a Segmentation Prompt Generation module which employs class-specific generators, isolating the prompt generation process for each class. This design approach confers significant architectural advantages when compared to conventional CSS methods. Traditional CSS approaches typically formulate segmentation as a pixel-level multi-class classification problem, wherein decision boundaries must be continuously calibrated to accommodate previously learned classes. Such recalibration inevitably leads to \textbf{error accumulation} across sequential tasks. By decoupling the prompt generation process for each class, we effectively isolate the learning of class-specific features, preventing interference between classes during continual learning.



\subsubsection{Channel Selection} 
The Channel Selector organizes the selected visual embeddings, $\mathbf{V}_i^{\text{sel}}$, based on their semantic associations, $\mathbf{C}_i$ (defined previously).  For each class $k$, it extracts the token embeddings most strongly associated with that class.  Formally, the class-specific token set, $\mathcal{T}_k$, is defined as \(\mathcal{T}_k=\{\mathbf{V}_i^{\text{sel}}[n,:]|(n,j)\in\mathbf{C}_i, j=k\}\).








\subsubsection{Prompt Generation (pGen)} 
To create a consistent input dimension for our prompt generator, we process these class-specific tokens as follows using a max token length $Q_{m}$ for every class, i.e., \(\hat{\mathbf{z}}_k = \text{Flatten}(\mathcal{T}_k) \oplus \mathbf{L}_k\), 
where $\text{Flatten}(\cdot)$ concatenates all tokens in $\mathcal{T}_k$ into a single vector and then adjusts the length to a fixed size \(Q_m\) by either padding with zeros or truncating.
$\mathbf{L}_k$ is a learnable class-specific embedding, and $\oplus$ denotes element-wise addition.
The resulting vector $\hat{\mathbf{z}}_k$ has a fixed dimension $d_z$. By generating prompts specific to each class, \textit{pGen could capture the unique segmentation characteristics of different classes}, which often exhibit distinct mask patterns and boundaries. This tailored prompt generation enhances the decoder's performance, enabling more precise segmentation across diverse object categories.
The final step involves transforming this class-specific representation into positional prompts for SAM. 
This is formulated as $\mathbf{p}_k = \text{pGen}_k(\hat{\mathbf{z}}_k)$, where $\text{pGen}_k$ is a class-specific two-layer MLP dedicated to class $k$, and $\mathbf{P}_k \in \mathbb{R}^{m \times d_p}$ represents $m$ prompt tokens with dimension $d_p$ (in our implementation, $m=6$).



\begin{table*}[t]
    \centering
    \caption[voc-bench]{Comparison with existing  methods on PASCAL VOC2012 in mIoU (\%). The 1\textsuperscript{st} highest results among Replay methods and the 1\textsuperscript{st} highest results among Data-free methods (excluding our method) are marked with \underline{underline}. $\dagger$ means the latest methods proposed from 2024 to 2025. * is on behalf of our own implementation.}
    \label{tab:voc}
   \resizebox{0.88\textwidth}{!}
    {
       \begin{tabular}{l|l|ccc|ccc|ccc|ccc}
        \toprule
     &\multirow{2}{*}{\textbf{Method}}                    & \multicolumn{3}{c}{\textbf{19-1} (2 tasks)}    & \multicolumn{3}{c}{\textbf{15-5} (2 tasks)}      & \multicolumn{3}{c}{\textbf{15-1} (6 tasks)}      & \multicolumn{3}{c}{\textbf{10-1} (11 tasks)} \\
                                       &                  & \textit{0-19} & \textit{20} & \textit{all}     & \textit{0-15} & \textit{16-20} & \textit{all}    & \textit{0-15} & \textit{16-20} & \textit{all}    & \textit{0-10} & \textit{11-20} & \textit{all} \\
                                        \midrule

    \multirow{5}{*}{\rotatebox[origin=c]{90}{Replay}}
                                        & MicroSeg-M~\cite{microseg}                              & --- & --- & --- & 82.9 & 60.1 & 77.5 & 82.0 & 47.3 & 73.3 & 78.9 & 59.2 & 70.1 \\
                                        &SSUL-M~\cite{samrep2021ssul}                             & \underline{77.38} & 22.43 & \underline{74.76} & 79.3 & 55.1 & 73.5 & 78.8 & 49.7 & 71.9 & 75.3 & 54.1 & 65.2 \\
                                        &RECALL~\cite{samrep2021RECALL}                           & 67.9 & \underline{53.5} & 68.4 & 66.6  & 50.9 & 64.0 & 65.7 & 47.8 & 62.7 & 59.5 & 46.7 & 54.8\\
                                        &SATS-M~\cite{SATS}                                       & --- & --- & --- &  81.44 & 70.02 & 78.72 & 80.37 & 64.54 & 76.61 &76.21 &61.62 &69.27\\
                                        & IPSeg-M$\dagger$~\cite{ipseg}                                    & --- & --- & --- & \underline{83.3} & \underline{73.3} & \underline{80.9} & \underline{83.5} & \underline{75.1} & \underline{81.5} & \underline{80.3} & \underline{76.7} & \underline{78.6} \\
                                        
                                        \midrule

    \multirow{12}{*}{\rotatebox[origin=c]{90}{Data-free}}  
                                        & MiB*~\cite{CVPR2020ModelingTB}                             & 69.91 & 20.63 & 67.45 & 75.48 & 49.41 & 68.47 & 36.71 & 12.12 & 30.82 & 12.20 & 13.19 & 12.61 \\
                                        & PLOP*~\cite{douillardPLOPLearningForgetting2021}                             & 74.15 & 35.58 & 72.04 & 75.49 & 49.66 & 69.34 & 64.09 & 20.12 & 53.11 & 44.03 & 15.51 & 30.45 \\
                                        & MiB+NeST$\dagger$*~\cite{ECCV2024EarlyPP}                          & 70.25 & 26.06 & 68.91 & 75.46 & 48.68 & 69.47 & 60.24 & 19.97 & 48.97 & 52.36 & 21.07 & 37.41 \\
                                        & PLOP+NeST$\dagger$*~\cite{ECCV2024EarlyPP}                       & 76.09 & 47.93 & 73.82 & 76.11 & 48.47 & 68.44 & 48.97 & 23.28 & 48.18  & 54.21 & 17.83 & 36.91\\
                                        & MicroSeg~\cite{microseg}                    & --- & --- & --- & 81.9 & 54.0 & 75.2 & 80.5 & 40.8 & 71.0 & 73.5 & 53.0 & 63.8 \\
                                        & CoMFormer~\cite{cermelliCoMFormerContinualLearning2023}                                       & 75.35 & 24.06 & 72.91 & 74.68 & 54.30 & 71.12 & 70.78 & 32.24 & 61.60 &  --- & --- & ---\\
                                        & CoMasTRe$\dagger$~\cite{CoMasTRe2024}                          & 75.13 & \underline{69.51} & \underline{74.86} & 79.73 & 51.93 & 73.11 & 69.77 & 43.62 & 63.54 &  --- & --- & ---\\
                                        & IPSeg~\cite{ipseg}                      & --- & --- & --- & \underline{81.4} & \underline{62.4} & \underline{76.9} & \underline{82.4} & 52.9 & \underline{75.4} & \underline{80.0} & \underline{61.2} & \underline{71.0} \\
                                        & SATS*~\cite{SATS}                                               &77.42 &61.07 &74.41 & 80.24 & 61.17 & 75.70 & 78.38 & \underline{62.02} & 74.48 & 64.27 & 58.66 & 61.60\\
                                        & BARM$\dagger$*~\cite{ECCV2024BackgroundAW}                             & \underline{77.6} & 41.4 & 75.2 & --- & --- & --- & 77.3 & 45.8 & 69.8 & 72.2 & 49.8 & 61.9 \\
                                        & SSUL~\cite{samrep2021ssul}                          & --- & --- & --- & 79.7 & 55.3 & 73.9 & 78.1 & 33.4 & 67.5 & 74.3 & 51.0 & 63.2 \\
                                         \rowcolor{gray!30}
                                        & \textbf{Ours}                    & \textbf{82.92} & \textbf{83.71} & \textbf{83.95} & \textbf{84.03} &  \textbf{81.68} & \textbf{83.47} & \textbf{83.81} & \textbf{82.12} & \textbf{83.40} & \textbf{84.03} &  \textbf{82.12} & \textbf{83.12} \\
    \bottomrule
\end{tabular}
    }
        
\end{table*}

\subsection{Class-Agnostic Segmentation and Aggregation}
\label{sec:cas}


The Class-Agnostic Segmentation (CAS) module utilizes a segmentation foundation model, and we currently employ SAM (Segment Anything Model)~\cite{kirillov2023segment} for this role. SAM works in a class-agnostic manner and can generate high-quality segmentation results for the input image solely based on spatial prompt conditions. Specifically, for each class-specific prompt $\mathbf{p}_k$ (from above section), 
SAM outputs a corresponding mask $\mathcal{M}_k$ and a confidence score. 
These individual class-specific masks are aggregated into a final semantic segmentation map. 




It is worth noting that the CAS module is frozen in the whole continual learning process. However, when the model learns task $t$, the outputs from CAS are used to generate the segmentation loss function. In our framework, the same segmentation loss as that for SAM training is adopted to train the learnable class-specific prompt generators, and the loss is also used together with the asymmetric loss~\cite{ridnik2021asymmetric} to train the task-specific LoRAs  in the LTCD module.

During model inference, given a test image, the task-specific and class-specific components of each task are plugged into the segmentation system to segment image regions specifically for those classes estimated appearing in the input image. Such inference process is run over all tasks and the results are collected for the final segmentation. The final segmentation is created by aggregating individual class masks. While pixel ownership is unambiguous for the vast majority of pixels, any rare spatial overlaps are resolved using a confidence-based approach. The pixel is assigned to the class with the highest confidence score from the SAM decoder.
Notably, this process requires no task ID, with time linear complexity with respect to the number of learned tasks $T$. Since only one adapter resides in memory at any time, maximum memory usage remains identical to that in single-task models. For time-sensitive applications, memory-time tradeoff techniques~\cite{s-lora} can be used to reduce inference latency.

\begin{table*}[!hbt]
    \centering
    \caption[ade20k-bench]{Comparison with existing methods on ADE20K in mIoU (\%). The 1\textsuperscript{st} highest results among Replay methods and the 1\textsuperscript{st} highest results among Data-free methods (excluding our method) are marked with \underline{underline}. $\dagger$ means the latest methods proposed from 2024 to 2025. * is on behalf of our own implementation.}
    \label{tab:ade}
     \resizebox{0.88\textwidth}{!}{
    \begin{tabular}{l|l|ccc|ccc|ccc}
    \toprule
    &\multirow{2}{*}{\textbf{Method}} & \multicolumn{3}{c}{\textbf{100-50} (2 tasks)}    & \multicolumn{3}{c}{\textbf{100-10} (6 tasks)}    & \multicolumn{3}{c}{\textbf{100-5} (11 tasks)}     \\
                     &                & \textit{0-100} & \textit{101-150} & \textit{all} & \textit{0-100} & \textit{101-150} & \textit{all} & \textit{0-100} & \textit{101-150} & \textit{all}  \\
    \midrule
    \multirow{3}{*}{\rotatebox[origin=c]{90}{Replay}}
                                            &IPSeg-M$\dagger$~\cite{ipseg}            & \underline{43.8} & 31.5 & \underline{39.7} & \underline{43.0} & \underline{30.9} & \underline{39.0} & \underline{43.2} & \underline{30.4} & \underline{38.9} \\
                                            &SSUL-M~\cite{samrep2021ssul}   & 41.5 & \underline{48.0} & 33.7 & 41.6 & 19.9 & 34.4 & 41.6 & 20.1 & 34.5\\
                                            &TIKP$\dagger$~\cite{yu2024tikp}     & 42.2 & 20.2 & 34.9 & 41.0 & 19.6 & 33.8 & 37.5 & 17.6 & 30.9 \\
    \midrule

    \multirow{11}{*}{\rotatebox[origin=c]{90}{Data-free}}
                                            & MiB*~\cite{CVPR2020ModelingTB}                             & 39.02 & 16.73 &   31.29  & 36.68 & 9.81 & 27.77  & 34.22 & 5.26 & 24.29  \\
                                            & PLOP*~\cite{douillardPLOPLearningForgetting2021}                             & 40.38 & 13.41 & 31.52  & 39.46 & 12.58 & 30.10 & 38.12 & 7.32 & 27.39 \\
                                            & MiB+NeST$\dagger$*~\cite{ECCV2024EarlyPP}                          & 38.84 & 23.11 & 33.55 & 38.79 & 19.10 & 32.24 & 38.39 & 17.46 & 31.23 \\
                                            & PLOP+NeST$\dagger$*~\cite{ECCV2024EarlyPP}                         & 40.78 & 22.78 & 34.84 & 39.42 & 20.50 & 33.21 & 37.83 & 16.89 & 30.53\\
                                            & CoMFormer~\cite{cermelliCoMFormerContinualLearning2023}                         & 44.70 & 26.20 & 38.40 & 40.60 & 15.60 & 32.30 & 39.50 & 13.60 & 30.90\\
                                            & CoMasTRe$\dagger$~\cite{CoMasTRe2024}                         & 45.73 & 26.02 & 39.20 & 42.32 & 18.42 & 34.41 & 40.82 & 15.83 & 32.55\\
                                            & BARM$\dagger$*~\cite{ECCV2024BackgroundAW}                              & 42.0 & 23.0 & 35.7 & 41.1 & 23.1 & 35.2 & 40.5 & 21.2 & 34.1 \\
                                            & BalConpas-R$\dagger$~\cite{ECCV2024StrikeStrikeaBalanceinContinualPanopticSegmentation}                        & \underline{50.8} & \underline{30.4} & \underline{44.0} & \underline{48.1} & 25.3 & \underline{40.5} & \underline{43.9} & 22.7 & 36.9 \\
                                            & IPSeg$\dagger$~\cite{ipseg}  & 43.2 & 29.0 & 38.4 & 42.5 & \underline{27.8} & 37.6 & 43.1 & \underline{26.2} & \underline{37.6} \\
                                            & SSUL~\cite{samrep2021ssul}   &41.9 & 20.1 & 34.6 & 40.7 & 19.0 & 33.5 & 41.3 & 16.0 & 32.9\\
                                             \rowcolor{gray!30}
                                            & \textbf{Ours}                    & \textbf{57.72} & \textbf{51.21} & \textbf{56.80} & \textbf{58.19} & \textbf{52.03} & \textbf{56.92} & \textbf{57.53} & \textbf{55.62} & \textbf{56.89} \\
    \bottomrule
\end{tabular}
}
\end{table*}

\section{Experiments}
\label{sec:exp}

\subsection{Experimental Setup}

\noindent\textbf{Datasets} Following previous studies~\cite{CVPR2020ModelingTB,douillardPLOPLearningForgetting2021}, we evaluate our method on PASCAL VOC2012~\cite{everinghamPascalVisualObject2010} and ADE20K~\cite{zhouSceneParsingADE20K2017}. PASCAL VOC2012 contains 20 object classes while ADE20K presents a more challenging scenario with 150 classes. Dataset statistics are in Appendix~\ref{sec:dataset}.

\noindent\textbf{CSS Settings.} The training process is divided into $T$ tasks with certain protocol. The \textit{overlapped} protocol, which is more challenging and realistic compared to the \textit{disjoint} setting, allows images to contain old classes previously learned and future classes to be learned~\cite{SATS}. For these images, annotations are provided only for the current task's classes and image regions corresponding to both old and future classes are annotated as background. Our experiments use this \textit{overlapped} setting throughout. For instance, in the VOC2012 10-1 (11 tasks) setting, the model first learns to segment 10 classes, then incrementally learns one new class in each of  10 new tasks. 

\noindent\textbf{Implementation details.} The model was trained for 5 epochs on PASCAL VOC2012 and 20 epochs on ADE20K using AdamW~\cite{loshchilov2017decoupled} optimizer (initial learning rate $1 \times 10^{-4}$, weight decay 0.05) with a polynomial learning rate schedule. For the LTCD module, we set the similarity threshold $\tau$ to 0.3 for PASCAL VOC2012 and 0.2 for ADE20K, ensuring an appropriate quantity of selected visual tokens in $\mathbf{V}_i^{\text{sel}}$, with $M=30$ descriptive phrases per class and LoRA rank $L=32$. The SPG module uses $m=6$ prompt tokens with a single hidden layer, and pGen input lengths of 512 and 1024 for PASCAL VOC2012 and ADE20K respectively, with the longer length for ADE20K accommodating its complex scene composition and higher object density. All experiments were conducted with batch size 24 on 8 NVIDIA 3090 GPUs.


\subsection{Main Results}
\begin{figure*}[!t]
    \centering
    \includegraphics[width=0.92\textwidth]{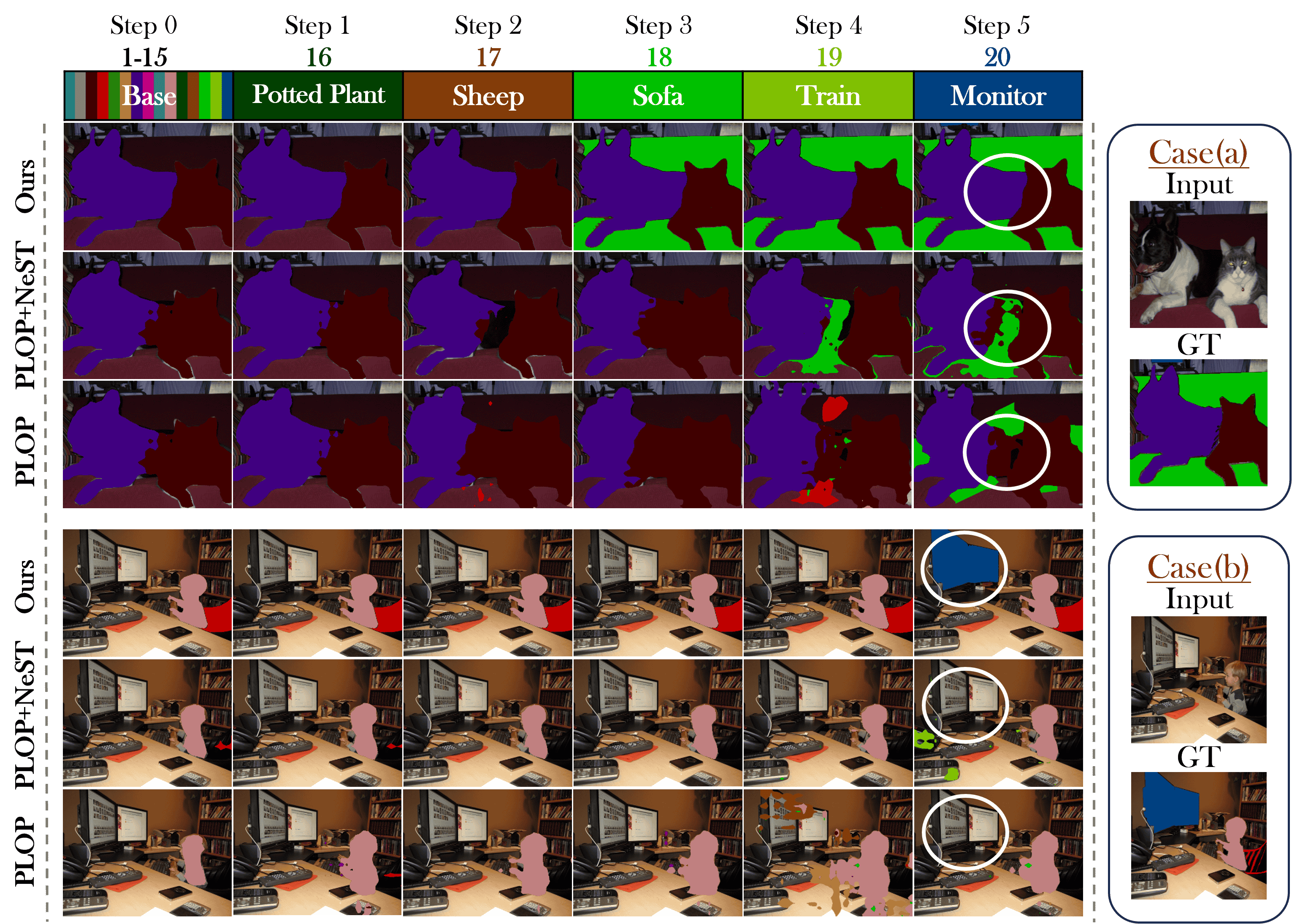}
    \caption{The visualization comparison from the last task on the 
Pascal VOC 2012 15-1 setting. We demonstrate our ability to resist forgetting and learn new categories}
    \label{fig:comparisonResults}
\end{figure*}

\cref{tab:voc,tab:ade} present the evaluation results of the proposed 
method on PASCAL VOC2012 and ADE20K. On VOC2012, our method achieves the best performance across all the four common CSS splitting settings, with 83.12\% in the challenging 10-1 setting, surpassing the previous SOTA method IPSeg (using data replay strategy) by a significant margin of 4.52\%. The visual examples in~\cref{fig:main_results} show that our method has good mask quality while maintaining resilience to semantic shift and catastrophic forgetting. 
It can be clearly seen from~\cref{fig:comparisonResults} that the existing methods have obvious limitations in the process of incremental learning. It is particularly noteworthy that these methods, in the last few steps of the Pascal VOC 2012 15-1 setup, are primarily concerned with preventing forgetting of previously learned categories rather than effectively learning newly introduced categories.
As the figure shows, other methods(PLOP and PLOP NeST) do not perform well in learning new categories ( such as TV Monitor for case b ). This is a direct reflection of their design and training philosophy - too much emphasis on preventing catastrophic forgetting and too little on the effective absorption of new knowledge. In case a, we can see method misclassify sofa into previously learned categories after task 4. Our proposed method can not only maintain accurate recognition of previous categories, but also efficiently learn and segment newly introduced categories.
Similarly on the more challenging ADE20K dataset, our method also excels in all the 100-50, 100-10 and 100-5 settings.  
Notably, our method significantly outperforms the SOTA method with a 17.99\% mIoU increase in the 100-5 setting.
These results demonstrate its strong performance in learning new knowledge while preserving old knowledge.  


Furthermore, in the more challenging CSS settings where the number of classes learned in the first task is the same as that in each subsequent task, including the settings 2-2 (10 tasks), 4-2 (9 tasks), and 4-4 (5 tasks), the superiority of our method becomes more evident. As shown in \cref{fig:combined}b and \cref{tab:voc_2-2},  our method consistently maintained both plasticity and stability in these highly challenging scenarios, outperforming competing methods by huge gaps of 74.94\%, 73.35\%, and 57.46\% respectively.

\begin{table}[!t]
    \centering
    \caption[voc-bench]{Quantitative comparison of different methods for challenging settings: 2-2, 4-2, and 4-4 incremental segmentation scenarios on PASCAL VOC 2012.}
    \label{tab:voc_2-2}
    \vspace{-2.0mm}
    \resizebox{\columnwidth}{!}{
       \begin{tabular}{ll|ccc|ccc|ccc}
    \toprule
     &\multirow{2}{*}{\textbf{Method}} & \multicolumn{3}{c}{\textbf{2-2} (10 tasks)} & \multicolumn{3}{c}{\textbf{4-2} (9 tasks)} & \multicolumn{3}{c}{\textbf{4-4} (5 tasks)} \\
    &                                   & \textit{0-2} & \textit{3-20} & \textit{all}  & \textit{0-4} & \textit{5-20} & \textit{all} & \textit{0-4} & \textit{5-20} & \textit{all} \\
    \midrule
                                         & MiB~\cite{CVPR2020ModelingTB}                              & 3.27 &   6.35 & 5.91 & 7.08 & 6.72 & 6.80 & 39.50 & 20.07 & 24.70 \\
                                          & PLOP~\cite{douillardPLOPLearningForgetting2021}                              & 24.96 & 5.38  & 8.18 & 15.41 & 6.55 & 8.67 & 28.54 & 15.95 & 18.95 \\
                                          & MiB+NeST~\cite{ECCV2024EarlyPP}                         & 13.44 & 6.72 & 7.68 & 10.09 & 7.51 &  7.77& 38.31 & 21.3 & 25.35\\
                                          & PLOP+NeST~\cite{ECCV2024EarlyPP}                        & 27.98 & 6.53 & 8.26 & 17.57 & 7.84 & 10.15 & 33.89 & 17.84 & 21.66  \\
    \midrule
    \rowcolor{gray!30}
    & \textbf{Ours}                    & \textbf{87.31} & \textbf{82.52} & \textbf{83.20} & \textbf{87.63} &  \textbf{82.22} & \textbf{83.50} & \textbf{87.63} & \textbf{81.31} & \textbf{82.81}  \\
    \bottomrule
\end{tabular}
    }
   \vspace{-3mm}
\end{table}

\begin{table}[!t]
    \centering 
    \caption{Ablation study on PASCAL VOC2012, evaluated in terms of mIoU (\%). The components include LoRA (L), Class-specific pGen (CG), and semantic aggregation (S).}\label{tab:ablation_component}    
    \resizebox{\columnwidth}{!}{
        \begin{tabular}{ccc|ccc|ccc|ccc}
        \toprule
        \multicolumn{3}{c|}{\textbf{Components}} & \multicolumn{3}{c|}{\textbf{19-1} (2 tasks)} & \multicolumn{3}{c|}{\textbf{10-1} (11 tasks)} & \multicolumn{3}{c}{\textbf{15-1} (6 tasks)} \\
        L & CG & S & \textit{0-19} & \textit{20} & \textit{all} & \textit{0-10} & \textit{11-20} & \textit{all} & \textit{0-15} & \textit{16-20} & \textit{all} \\
        \midrule
        \xmark & \xmark & \xmark & 25.16 & 62.79 & 26.95 & 22.32 & 25.81 & 23.98 & 25.42 & 26.28 & 25.62 \\
        \cmark & \cmark  &\xmark & 79.37 & 80.26 & 79.41 & 81.22 & 80.51 & 80.88 & 80.35 & 80.82 & 80.46 \\
        \cmark & \xmark  &\cmark & 29.54 & 77.41 & 31.81 & 28.76 & 27.43 & 27.89 & 30.14 & 28.17 & 28.83 \\
        \xmark & \cmark  &\cmark & 76.13 & 77.97 & 76.21 & 77.64 & 76.33 & 77.01 & 78.38 & 77.12 & 78.03 \\
        \cmark & \cmark  &\cmark & 82.92 & 83.71 & 83.95 & 84.03 & 82.12 & 83.12 & 83.81 & 82.12 & 83.40 \\
        \bottomrule
        \end{tabular}
    }
   \vspace{-6mm}
\end{table}
\subsection{Ablation Studies}
Ablation studies were performed to verify the validity of 
three key components in our framework: LoRA, pGen, and semantic aggregation.
As shown in~\cref{tab:ablation_component}, removing each of three components leads to worse performance in all task settings. 
In addition, when class-specific pGen was replaced by classes-shared pGen, a dramatic drop particularly in 10-1 and 15-1 settings was observed. 
This drop happened because, with the shared pGen, the system had to adjust for each new class during training, as shown by the images in~\cref{fig:class-sharedMLP}. 
All these results confirm the necessity of the proposed components in our framework.




\vspace{-1.5mm}
\subsection{Sensitivity Studies}
\begin{figure}[t]
        \centering        
        \includegraphics[width=0.9\linewidth]{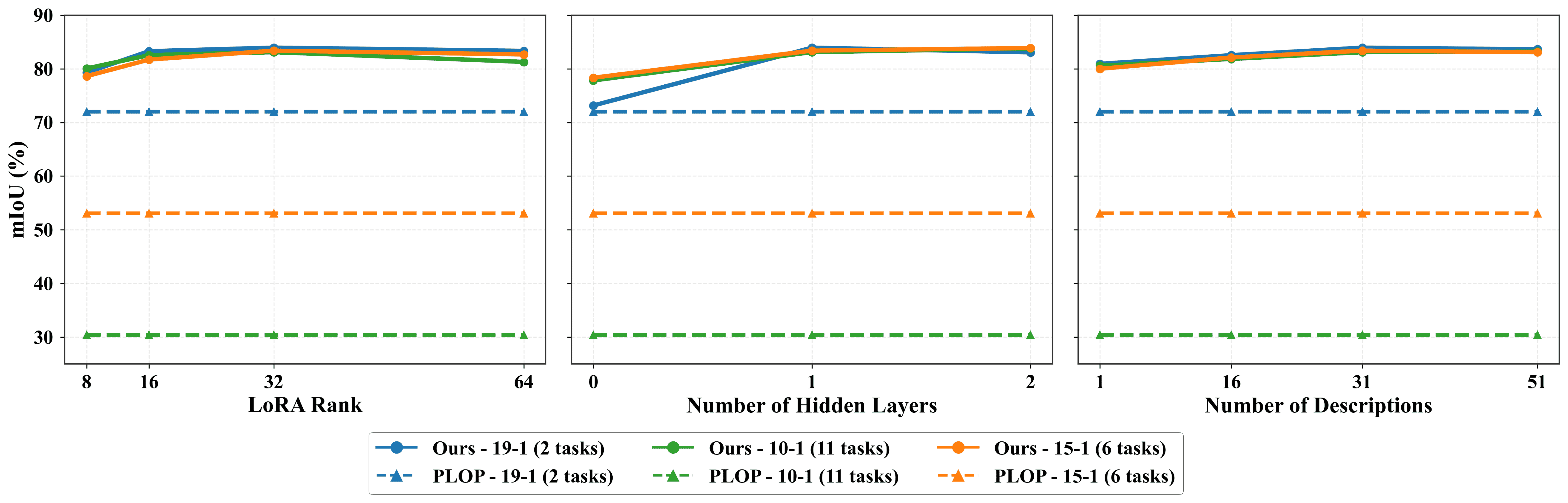}
        \vspace{-2mm}
        \caption{Sensitivity study 
        on PASCAL VOC2012.} 
        \label{fig:exp_sensitive}
\vspace{-4mm}
\end{figure}

The sensitivity of our method to hyper-parameter choice is also evaluated. As demonstrated in \cref{fig:exp_sensitive}, when varying the LoRA ranks within the range $[16, 64]$, the number of hidden layers in the class-specific pGen module within the range $[0, 2]$, the number of textual descriptions for each class based on the LLM within the range $[1, 51]$, our method performs stably well and always better than the representative strong baseline, supporting the robustness of our method to hyper-parameter choice.

 \begin{figure}[t]
    \centering
    \includegraphics[width=0.9\linewidth]{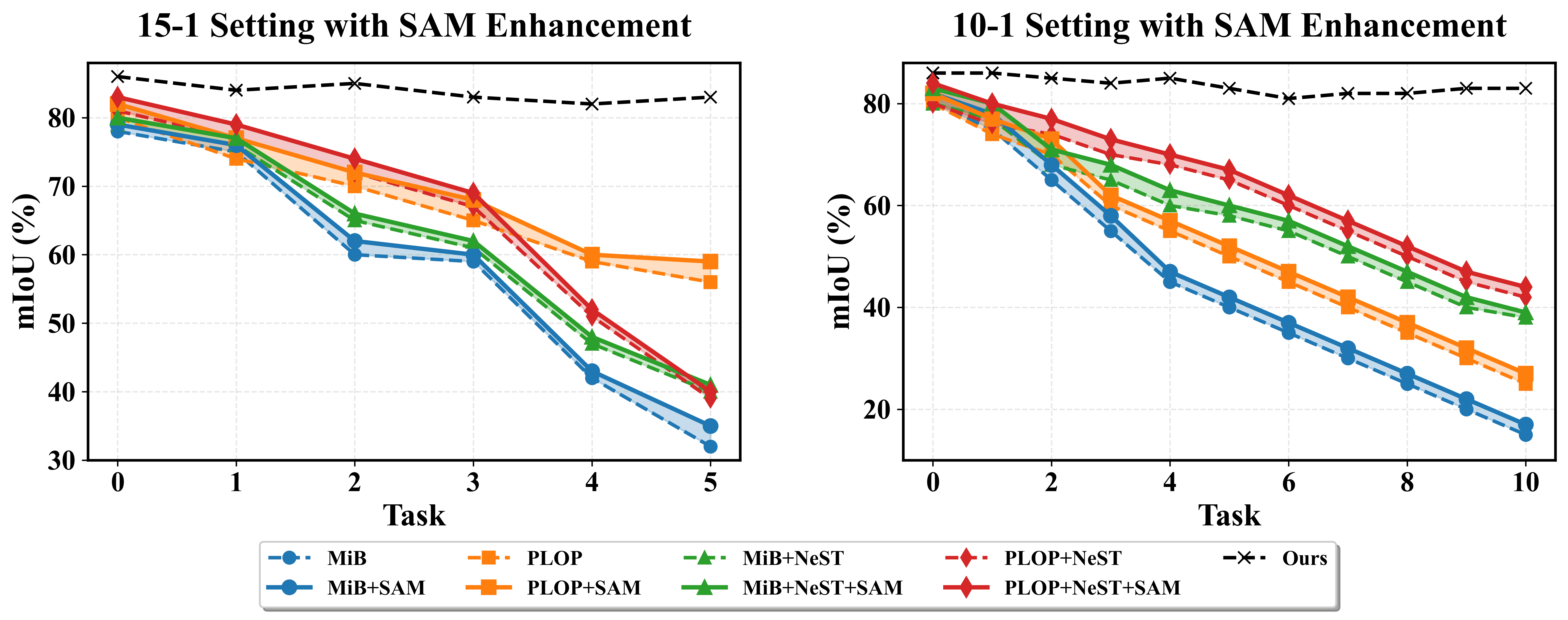}
    \caption{Segmentation performance on PASCAL VOC2012 with and without SAM enhancement.}
    \vspace{-2mm}
    \label{fig:method_with_sam}
\end{figure}
\vspace{-2mm}

\subsection{Additional Studies}
Additional experiments 
demonstrate that simply incorporating powerful foundation models like SAM is insufficient for effective CSS, and that the proposed class-specific prompt generation in our method is necessary.

As shown in~\cref{fig:method_with_sam}, simply applying SAM as a post-processing module to existing CSS methods yields only marginal improvements, with less than 6\% mIoU across various settings. This limited gain reveals fundamental limitations when SAM is applied directly to CSS outputs: blurred edges in CSS predictions compromise SAM's segmentation capabilities, and SAM's class-agnostic nature cannot correct semantic inconsistencies inherent in CSS outputs. In contrast, our method with the proposed prompt generation module which generates precise class-aware positional embeddings overcomes these limitations by intelligently guiding SAM to produce refined segmentation masks.

To further confirm the efficacy of the proposed prompt generation module, ground-truth class existence information and bounding box for each class region were first obtained for each test image by Grounding DINO. These boxes were then used as spatial prompts for SAM to produce segmentation masks.
Such approach achieves 84\% mIoU on PASCAL VOC2012 and 57\% mIoU on ADE20K. While this performance is comparable to our method (83\% on VOC2012 and 56\% on ADE20K), it is important to note that our approach achieves this without requiring ground-truth tags during inference and operates under the challenging constraints of continual learning. This demonstrates that our class-specific prompt generation approach effectively bridges detection and segmentation in CSS scenarios, achieving near-optimal performance even without perfect class knowledge.

Finally, we analyse the computational cost of our approach. Similar to previous continual learning studies which employ task-specific modules~\cite{Li_2025_CVPR}, our method introduces computational overhead during inference primarily from switching task-specific LoRA parameters. The inference time for a single task is 0.3 second per image, and the full process for six tasks in the (15-1) setting takes about 1.9 seconds per image. In comparison, most existing single encoder-decoder based CSS methods takes around 0.5 second per image. Despite leveraging foundation models, our parameter growth remains highly manageable. Each class requires only 8MB for the pGen module (0.225\% of total model size), while all tasks' LoRA adapters collectively need merely 22.18MB (0.62\% of total model size). Each task requires relatively few trainable parameters per increment~\cite{ji2023continual}, yet achieves state-of-the-art performance with remarkable gains in challenging settings, demonstrating practical applicability comparable to offline-trained models. Given our method's substantial performance gains and affordable storage overhead, this trade-off is well-justified for non-realtime applications.

\section{Conclusion}
\label{sec:conclusion}

In this work, we propose a two-stage framework for Continual Semantic Segmentation (CSS) that decouples class-aware detection from class-agnostic segmentation. Our method achieves state-of-the-art performance across various CSS tasks. The superior performance stems from task-specific modeling with conflict-free formulation.
By leveraging pre-trained text and image encoders, this study provides a practical solution for real-world applications such as autonomous driving and medical imaging, offering a scalable path for CSS.
The main limitation of our method is the inference time due to the sequential switching of task-specific parameters. 
Future work could explore parameter merging techniques to address this scalability issue.





{
    \small
    \bibliographystyle{ieeenat_fullname}
    \bibliography{main}
}

\appendix
\setcounter{equation}{0}
\renewcommand{\theequation}{A.\arabic{equation}}
\setcounter{page}{1}
\setcounter{table}{0}
\setcounter{figure}{0}
\setcounter{algorithm}{0}

\renewcommand{\thefigure}{\thesection\arabic{figure}}
\renewcommand{\thetable}{\thesection\arabic{table}}
\renewcommand{\thealgorithm}{\thesection\arabic{algorithm}}

\appendix
\setcounter{secnumdepth}{2}
\renewcommand{\thesection}{\Alph{section}}

\section{Additional Experimental Results}
\label{sec:additionalStudies}
\subsection{Additional Studies on Using SAM}
\label{appendix_sam_otherMethod}
Although SAM is a powerful model, our method's superior performance stems from how we integrate it rather than just its capabilities alone. We evaluated using fine-tuning SAM which was introduced in~\cref{sec:SAM_fine-tuning} to refine outputs from existing CSS methods. This limited gain reveals that merely applying SAM as a refinement module fails to harness its full potential, especially with CSS models whose performance degrades during incremental learning.

\begin{figure}[t]
    \centering
    
    \begin{subfigure}[b]{\linewidth}
        \includegraphics[width=\linewidth]{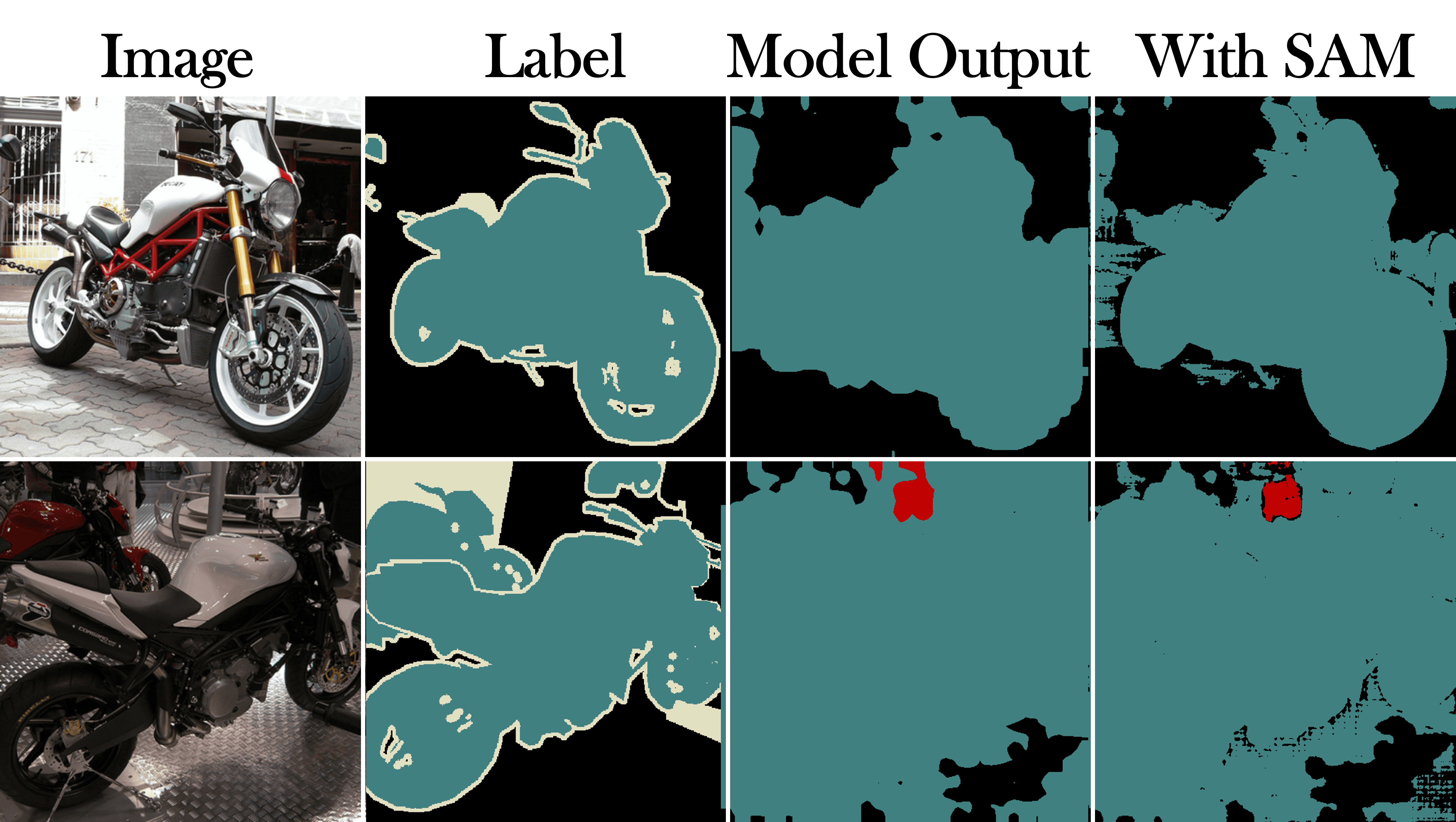}
        \caption{Output cannot be refined as expected due to blurred edges}
        \label{fig:refined_plop_1}
    \end{subfigure}
    \hfill
    \begin{subfigure}[b]{\linewidth}
        \includegraphics[width=\linewidth]{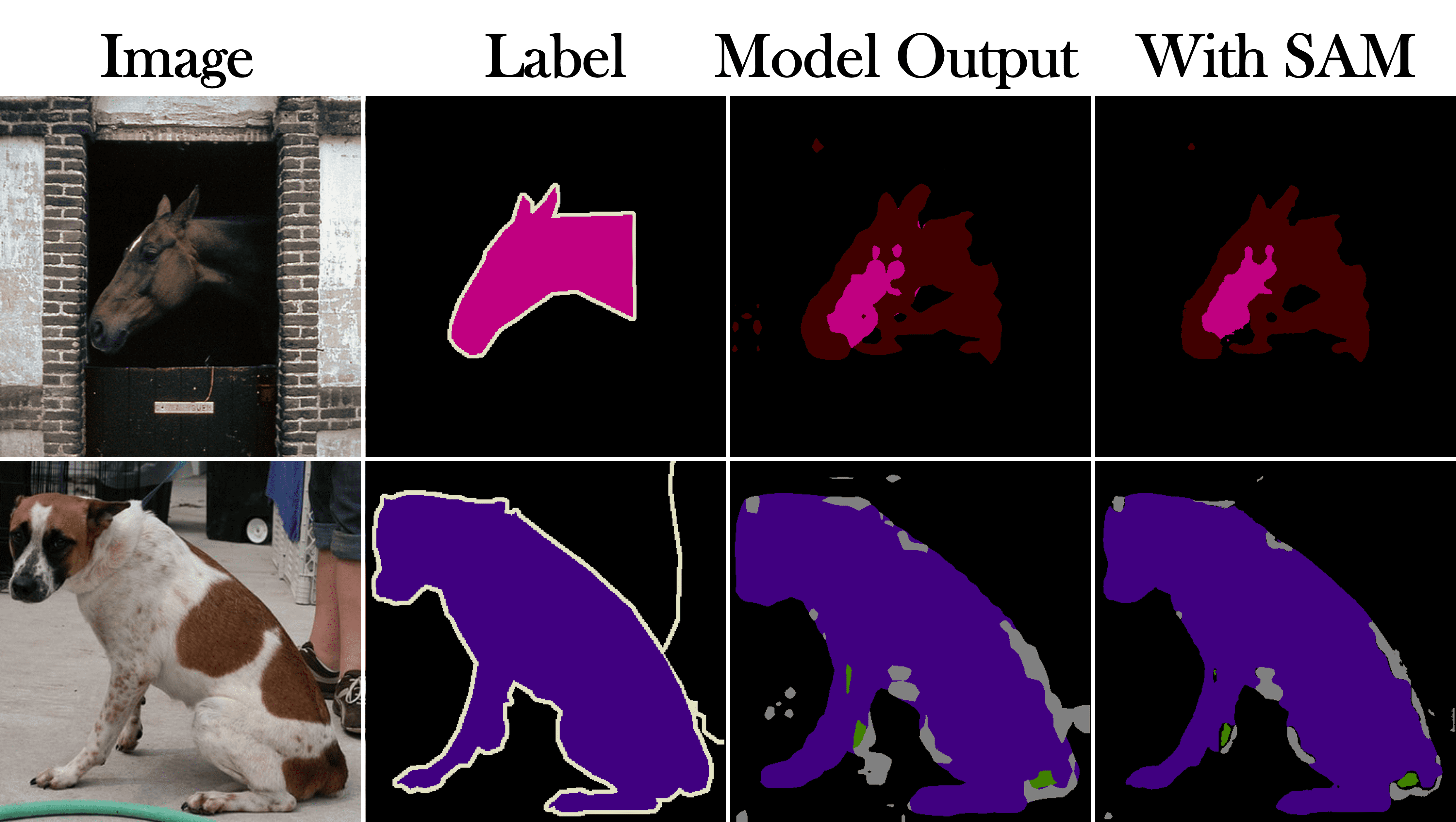}
        \caption{SAM fails to correct semantic shifts in the output}
        \label{fig:refined_plop_2}
    \end{subfigure}
    \caption{PLOP results on VOC with 10-1 task setting, showing limitations of SAM refinement due to semantic shifts.}
    \label{fig:refined_example}
\end{figure}

As~\cref{fig:refined_example} shows, we identified two key challenges when using SAM as a plug-and-play solution for CSS tasks:

\noindent\textbf{Blurred Edges.} Ambiguous boundaries in CSS predictions compromise SAM's effectiveness, as it relies on clear textural patterns for object delineation. When faced with blurred edges from CSS outputs, its extraction mechanism fails, as shown in \cref{fig:refined_plop_1}.

\noindent\textbf{Semantic Shift.} SAM's class-agnostic segmentation creates limitations in CSS contexts where semantic shifts occur. As illustrated in \cref{fig:refined_plop_2}, SAM cannot correct semantic inconsistencies without class-specific understanding.

To further validate our approach, we compared against a Grounding DINO with fine-tune SAM baseline, providing class information to Grounding DINO during inference. (detailed in \cref{sec:app_zeroshot}).
\subsection{Grounding DINO with SAM}
\label{sec:app_zeroshot}
In this section, we design a new benchmark method that combines Grounding DINO with the SAM model. Specifically, we provide accurate category information for each image, using the output of Grounding DINO as a prompt for SAM segmentation. 

\begin{figure}[h]
    \centering
    \begin{subfigure}[b]{0.45\linewidth}
        \includegraphics[width=\linewidth]{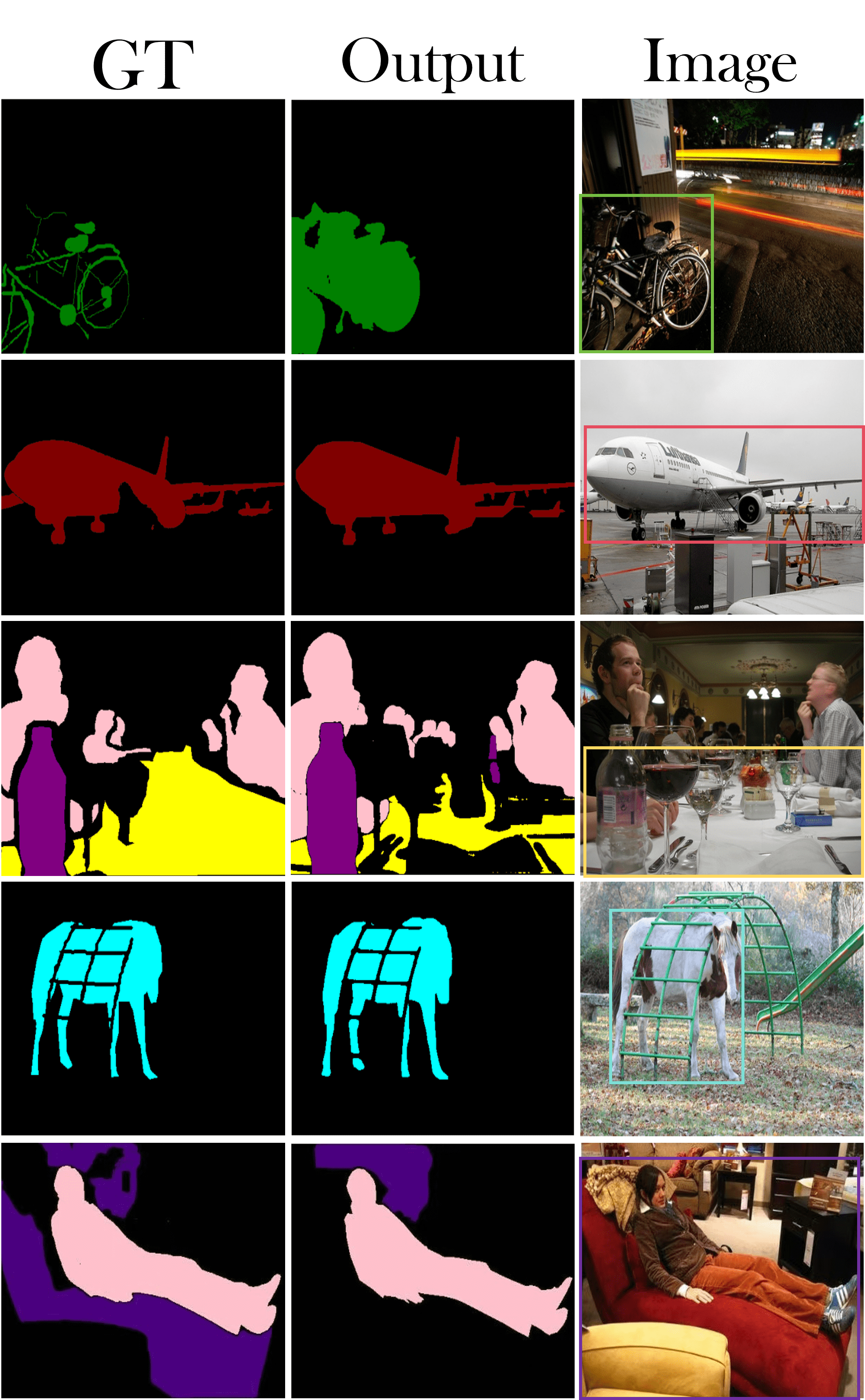}
        \caption{Test Grounding Dino with SAM in VOC}
        \label{fig:grounding_dino_voc}
    \end{subfigure}
    \hfill
    \begin{subfigure}[b]{0.45\linewidth}
        \includegraphics[width=\linewidth]{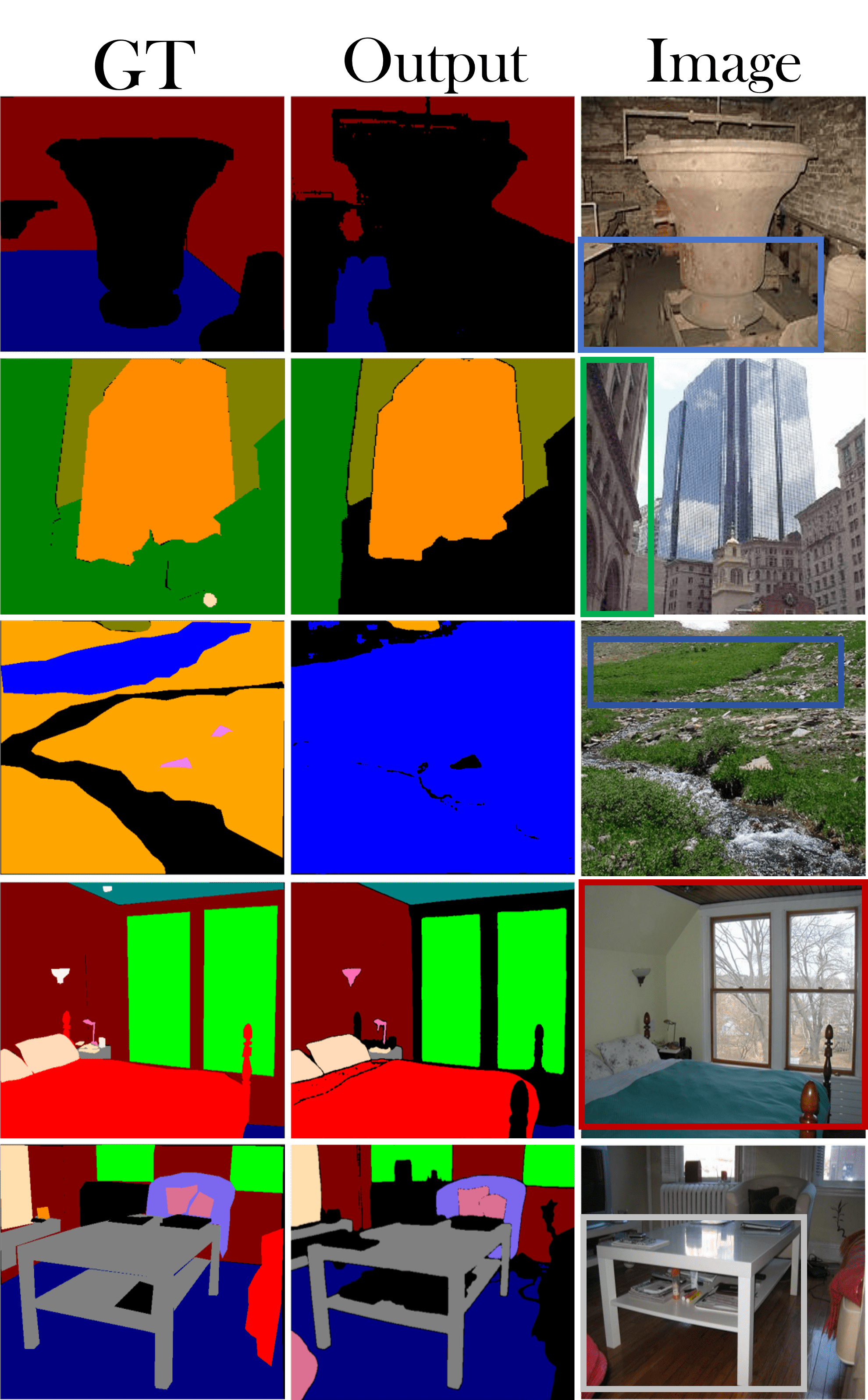}
        \caption{Testing Grounding Dino with SAM in ADE}
        \label{fig:grounding_dino_ade}
    \end{subfigure}
    \caption{the stars with red outline are background points, the stars with green outline are foreground points}
    \label{fig:prompt_example}
\end{figure}

On PASCAL VOC, the approach achieves 84\% mIoU. This excellent performance is due to the relatively simple structure of the dataset, in that images typically contain only 1-2 target categories. Grounding DINO generates accurate bounding boxes, allowing SAM to segment precisely with minimal supervision. However,as shown by~\cref{fig:grounding_dino_voc} line 5, SAM still struggles with suboptimal results due to the single-box prompt, as seen in ~\cref{fig:grounding_dino_voc}, which is more obvious in ADE dataset.

On the more complex ADE20K dataset, the benchmark's performance drops to 57\% mIoU. This is due to two main challenges: ADE20K’s multi-category nature complicates Grounding DINO’s ability to localize each class, and SAM’s reliance on bounding boxes leads to hallucinations—incorrect segmentation. These issues highlight the difficulties and limitations of applying simple connection with SAM and Grounding DINO to complex segmentation tasks.

\subsection{Parameters Analysis}
\label{sec:parameters_analysis}

We calculate the parameter size of our method in~\cref{tab:pgen_params}. The table presents the parameter statistics for a single class pGen across different datasets. For the VOC dataset, each class-specific pGen accepts an input length of 512 and contains approximately 2.1 million parameters, requiring only 8MB of storage space when using float32 precision. For the ADE dataset, the input length increases to 1024, resulting in about 2.6 million parameters and 10MB of storage.
\begin{table}[h]
    \centering
    \resizebox{\columnwidth}{!}{%
    \begin{tabular}{|c|c|c|c|}
        \hline
        \textbf{Datasets} & \textbf{Input length} & \textbf{Parameters} & \textbf{Size (MB)}\\
        \hline
        VOC & 512 & 2,097,152 & 8\\
        ADE & 1024 & 2,621,440 & 10\\
        \hline
    \end{tabular}%
    }
    \caption{pGen parameters for different datasets. The parameter size is calculated based on float32 (4 bytes per parameter).}
    \label{tab:pgen_params}
\end{table}

\subsection{Visualization Analysis}
\label{sec:visualization_analysis}
\begin{figure}[ht]
\label{fig:app_pGen}
   \centering
   \includegraphics[width=.8\linewidth]{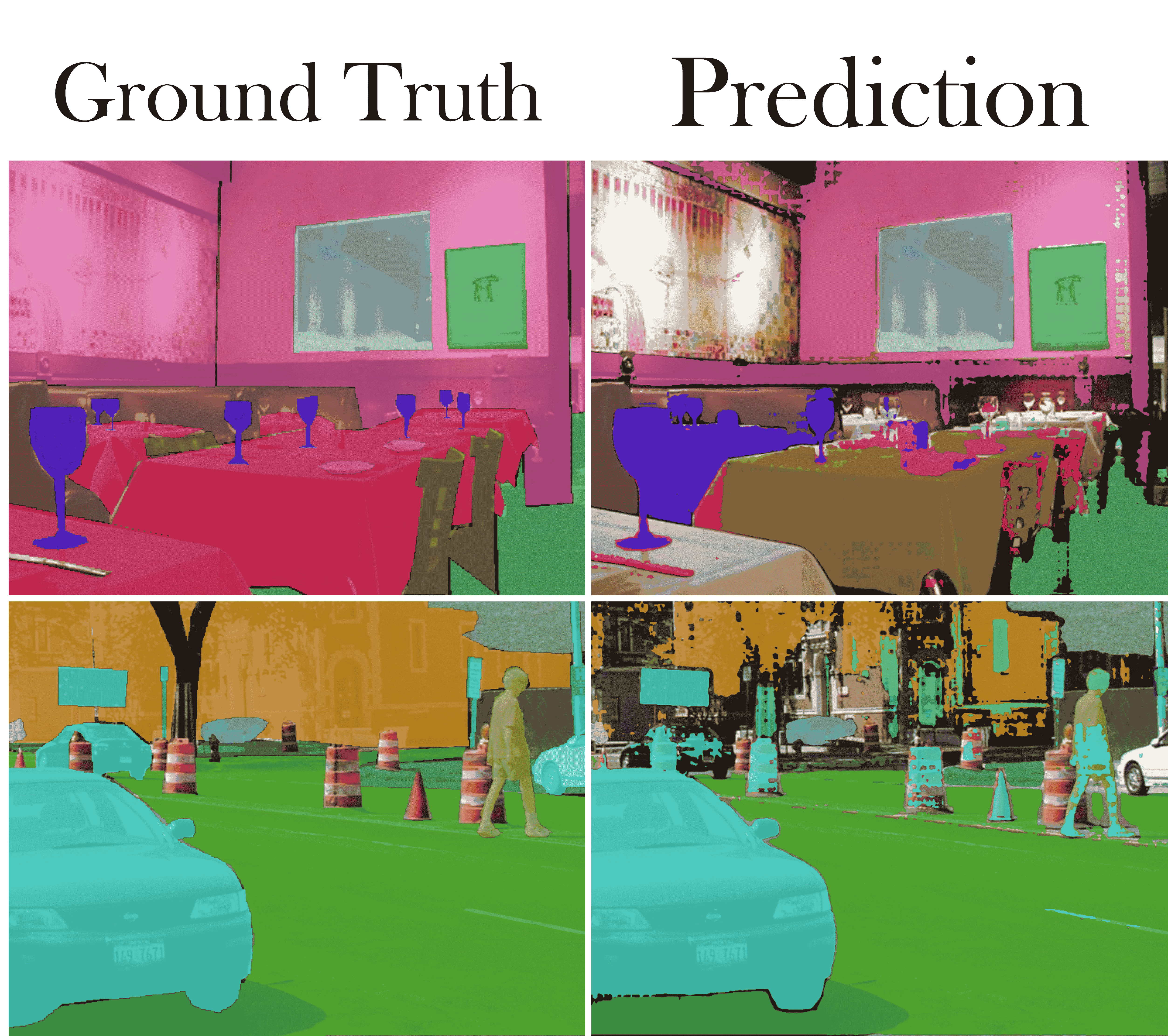}
   \caption{Visualization in ADE with class-shared pGen.Left: Semantic shift phenomenon observed. Right: Generation of imprecise positional embeddings leading to poor segmentation.}
   \label{fig:class-sharedMLP}
\end{figure}

pGen shared and class specific pGen, visulization in~\cref{fig:class-sharedMLP}.



\section{Implementation Details}
\label{sec:app_Implement_Detail}
\subsection{Description of Classes}
\label{sec:prompt}
The following prompts were employed to generate concise, adjective-based descriptive phrases for each class. Our approach leverages phrase-based descriptions rather than full sentences. This strategy is novel in that it constructs rich adjective+\{class name\} combinations to capture the essential visual features of each class. Such phrases have several CLASSes over full sentences: 
\begin{itemize}
    \item They reduce computational load by saving tokens.
    \item They eliminate unnecessary grammatical words, thus focusing on key visual semantics.
    \item They provide a more flexible and enriched representation in the visual-language space. For example, while the token “dog” may cluster only with generic animal features, a phrase like “red dog” can better capture subtle distinctions, facilitating improved matching with diverse image scenarios.
    \item They enhance the meaningfulness of self-attention mechanisms by avoiding extraneous words (e.g., “there is a”) that do little to describe the class.
\end{itemize}
The following prompts were used:
\begin{enumerate}[label=\arabic*.]
    \item ``Generate a concise descriptive phrase for a(n) \{CLASS\}."
    \item ``Provide a short adjective-based description for a(n) \{CLASS\}."
    \item ``What is a brief phrase that captures the visual essence of a(n) \{CLASS\}?"
    \item ``Compose a succinct phrase highlighting the key characteristics of a(n) \{CLASS\}."
    \item ``Summarize the appearance of a(n) \{CLASS\} using a short descriptive phrase."
    \item ``Formulate an adjective+\{class name\} combination that best describes a(n) \{CLASS\}."
    \item ``Identify a brief phrase that encapsulates the defining features of a(n) \{CLASS\}."
    \item ``Express the distinctive visual traits of a(n) \{CLASS\} in a concise phrase."
    \item ``What is the most representative adjective phrase for a(n) \{CLASS\}?"
    \item ``Create a short descriptive combination that emphasizes the visual attributes of a(n) \{CLASS\}."
    \item ``For classes with synonymous forms (e.g., `man', `woman', `girl' for person), generate a concise descriptive phrase considering these variations."
    \item ``Taking into account synonymous terms for \{CLASS\}, produce a succinct descriptive phrase capturing the core visual features."
\end{enumerate}
In our experiments, we employ GPT-4~\cite{openai2024gpt4technicalreport} and DeepSeek-V3~\cite{deepseekai2025deepseekv3technicalreport} to generate these phrases. An automated script is then used to filter and clean the outputs, ensuring the generated descriptions are both concise and semantically rich.

\subsection{Training Details}
\label{sec:training}
In training, we initialized our model with parameters from the Grounding DINO framework. Specifically, we leveraged the pre-trained image encoder, text encoder, and cross-attention component, with the latter referred to as the "Feature Enhancer" in the Grounding DINO paper. During the training process, we employed a task-specific LoRA strategy to fine-tune the cross-attention module while keeping both the image and text encoders frozen, which allows us to preserve the generalized feature extraction capabilities of the encoders while adapting the cross-modal interaction mechanism to our specific task requirements. We use CE loss function and Dice loss functions:~\cref{eq:total_loss}

During training, we adopted a multi-task learning strategy to optimize various model components. The primary loss function \(\mathcal{L}\) consists of a linear combination of cross-entropy (CE) and DICE losses, which jointly updates the parameters of the Class-specific pGen, task-specific LoRA adapters in cross-modal attention modules and the image encoder. 

This primary loss~\cref{eq:total_loss} design effectively guides the model to generate accurate segmentation masks aligned with ground truth labels. 
\begin{equation}
\begin{split}
\mathcal{L} &= \mathcal{L}_{\text{Dice}} + \mathcal{L}_{\text{CE}}\\[1mm]
&= 1 - \frac{2\sum_{i=1}^{N}p_i\,g_i}{\sum_{i=1}^{N}\left(p_i^2+g_i^2\right)}
-\frac{1}{N}\sum_{i=1}^{N}\sum_{c=1}^{C}y_{ic}\log(p_{ic}).
\end{split}
\label{eq:total_loss}
\end{equation}

Here, \(p_i\) denotes the predicted probability at the \(i\)-th pixel, \(g_i\) is the corresponding ground truth, \(y_{ic}\) is the one-hot encoded label for the \(i\)-th pixel for class \(c\), \(p_{ic}\) represents the predicted probability that the \(i\)-th pixel belongs to class \(c\), \(N\) is the total number of pixels, and \(C\) is the number of classes.

To derive classification results from the Similarity Matrix, we compute the average activation value across all image tokens for each text embedding. Specifically, the classifier output vector \(\mathbf{r}\) is calculated as:
\begin{equation}
\mathbf{r}[j] = \frac{\sum_{i=1}^{H} \mathbf{S}'_i[i,j] \cdot \mathbb{1}(\mathbf{S}'_i[i,j] > 0)}{\sum_{i=1}^{H} \mathbb{1}(\mathbf{S}'_i[i,j] > 0)}\quad j\in \{1,2,\cdots,c_t\}
\end{equation}
where $\mathbb{1}(\cdot)$ is the indicator function that equals 1 when the condition is true and 0 otherwise.

Simultaneously, we introduced Asymmetric Loss~\cite{ridnik2021asymmetric} as an auxiliary loss function~\cref{eq:asymmetric}, specifically designed to optimize the prediction $\mathbf{r}$ and image category classification results. This loss function is particularly suitable for handling class imbalance in multi-label classification problems, effectively adjusting the model's sensitivity to different categories. This dual-loss mechanism ensures that the model maintains category detection accuracy while generating high-quality location-aware prompts.

\begin{equation}
    \begin{split}
        ASL(p, y) = \begin{cases}
(1-p)^{\gamma_+} \log(p) & \text{if } y = 1 \\
p^{\gamma_-} \log(1-p) & \text{if } y = 0
\end{cases}
    \end{split}
    \label{eq:asymmetric}
\end{equation}


\section{Dataset}
\label{sec:dataset}
\noindent\textbf{Datasets. }Our method was evaluated on two semantic segmentation datasets 
PASCAL VOC2012~\cite{everinghamPascalVisualObject2010} and ADE20K~\cite{zhouSceneParsingADE20K2017}.
PASCAL VOC 2012 serves as a relatively straightforward benchmark containing 20 object classes plus a background class, containing 10,582 samples for training and 1,449 for validation, 
while ADE20K presents a more challenging scenario which is a large-scale semantic segmentation dataset with 150 annotated classes and one background class, containing 20,210 and 2,000 samples for training and validation.

\section{Fine-tuning SAM}
\label{sec:SAM_fine-tuning}
Due to the fact that different datasets have different styles and annotation standards, using a frozen mask decoder cannot satisfy the requirement for generating high IoU masks. To address this issue, we employ a vanilla fine-tuning method with learning rate decay.

We observe that SAM's segmentation performance under zero-shot conditions is significantly affected by dataset annotation styles, showing suboptimal performance on complex datasets. Therefore, we aim to fine-tune SAM to adapt to new data annotation patterns and improve its segmentation results.

The medical image segmentation field has presented many excellent examples and insights for fine-tuning SAM, where they apply LoRA to the Image Encoder while fine-tuning the Mask Decoder to adapt to their datasets. 

Through our research, we find that modifications to the Image Encoder in medical imaging, which yield remarkable results, are primarily necessary due to SAM's insufficient training on medical image acquisition. Since we are working in the CSS domain, which doesn't face these medical imaging-specific challenges, we reasonably hypothesize that fine-tuning only the \textbf{Mask Decoder} should be sufficient for SAM to adapt to CSS dataset(VOC,ADE).

We modify the semantic segmentation datasets to make them compatible with SAM's training framework. As shown in~\cref{fig:fintune_prompt_example}, we transform the original semantic segmentation labels into instance-level masks with corresponding prompts (e.g., point or box prompts), allowing SAM to utilize these datasets effectively during fine-tuning. We also create separated labels to maintain SAM's ability.


\subsection{Origin Label}

\begin{figure}[h]
    \centering
    \begin{subfigure}[b]{0.45\linewidth}
        \includegraphics[width=\linewidth]{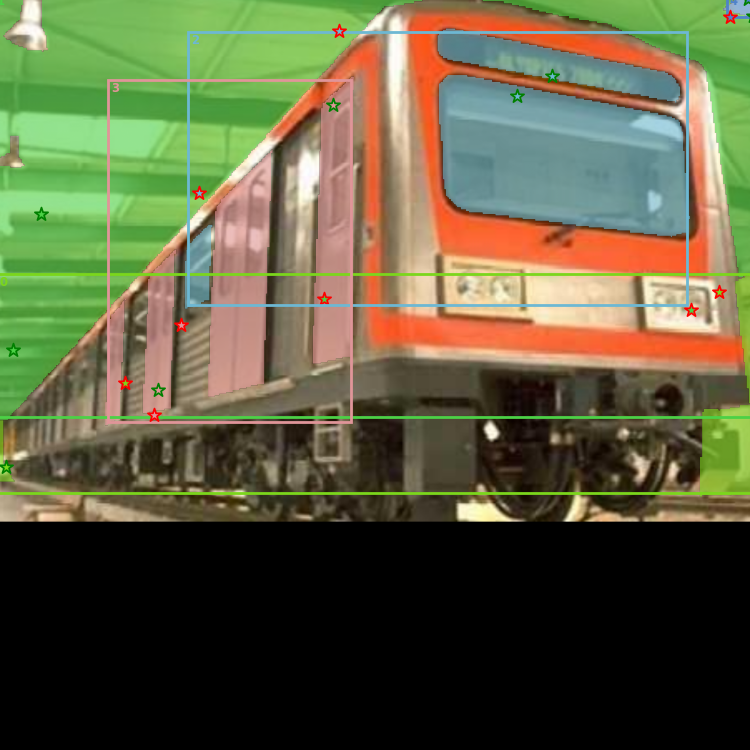}
        \caption{Origin label example}
        \label{fig:origin_label}
    \end{subfigure}
    \hfill
    \begin{subfigure}[b]{0.45\linewidth}
        \includegraphics[width=\linewidth]{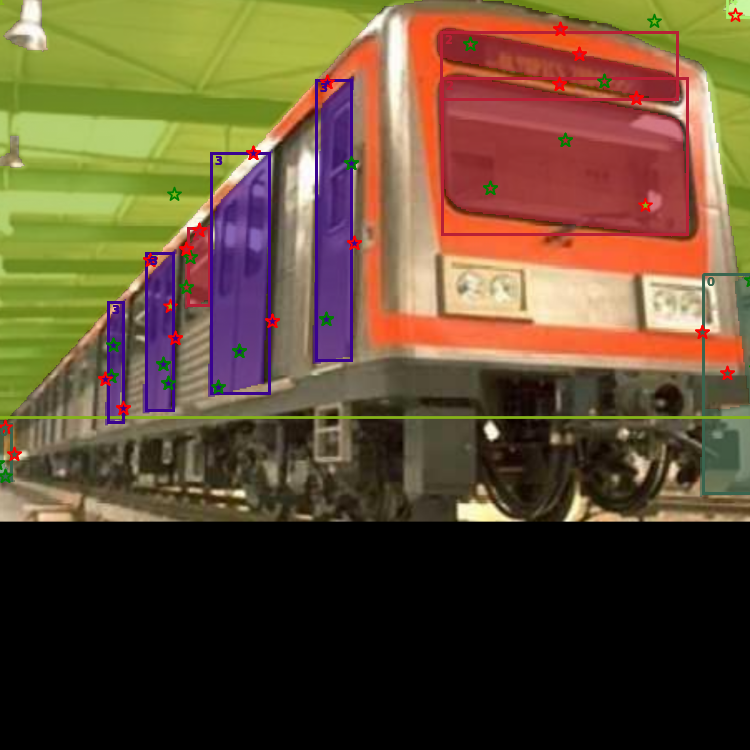}
        \caption{Separated label example}
        \label{fig:separated_label}
    \end{subfigure}
    \caption{the stars with red outline are background points, the stars with green outline are foreground points}
    \label{fig:fintune_prompt_example}
\end{figure}

In the ADE and VOC datasets, although it only provides segmentation labels without box or points, we can generate corresponding box prompts and point prompts from these labels. Point prompts can be divided into two categories: positive points placed on a specific label, and negative points placed within the box prompt but not on the corresponding label which is shown in~\cref{fig:origin_label}

\subsection{Separated Label}

With the aim to save grained ability for SAM, origin labels are separated based on whether they are in contact with other pixel with same label index, while excluding larger areas. We do the same prompt creating approach and example can be seen in~\cref{fig:separated_label}

\subsection{Fine-tuning Method in Mask Decoder}

This fine-tuning strategy enables the mask decoder to better adapt to different dataset characteristics while maintaining model generalization ability, resulting in more accurate segmentation results.
\vspace{3px}

\textbf{Parameter Update Strategy:}
\begin{itemize}
    \item We fine-tune part of parameters of the mask decoder(don't change the context embeddings)
    \item Use a small initial learning rate(1e-5) to maintain model stability 
    \item Gradually decrease the learning rate during training to avoid performance degradation from excessive updates
\end{itemize}



\subsection{Loss}  
Our loss(~\cref{eq:total_loss}) is a linear combination of Dice loss and cross-entropy (CE) loss to supervise the mask prediction, balancing region overlap and pixel-level classification accuracy.
\end{document}